\documentclass[sigconf,nonacm]{acmart}

\usepackage{tabularx}
\usepackage{multirow} 
\usepackage{subcaption}
\usepackage{booktabs}
\usepackage{makecell}
\usepackage{fancyhdr}
\AtBeginDocument{%
  }

\begin{document}

\title{Shallow Features Matter: Hierarchical Memory with Heterogeneous Interaction for Unsupervised Video Object Segmentation}

\thanks{\textcopyright{} [Xiangyu Zheng 2025] This is the author's version of the work. It is posted here for your personal use. Not for redistribution. The definitive version was published
in MM'25: The 33rd ACM International
Conference on Multimedia Proceedings, https://doi.org/10.1145/3746027.3755848.}

\author{Xiangyu Zheng}
\authornote{Equal contribution.}
\email{xyzheng23@m.fudan.edu.cn}
\affiliation{%
  \institution{Shanghai Key Lab of Intelligent Information Processing, College of Computer Science and Artificial Intelligence, Fudan University}
  \city{Shanghai}
  \country{China}
}
\author{Songcheng He}
\email{24210240166@m.fudan.edu.cn}
\authornotemark[1]
\affiliation{%
  \institution{Shanghai Key Lab of Intelligent Information Processing, College of Computer Science and Artificial Intelligence, Fudan University}
  \city{Shanghai}
  \country{China}
}
\author{Wanyun Li}
\email{wyli22@m.fudan.edu.cn}
\affiliation{%
  \institution{Shanghai Key Lab of Intelligent Information Processing, College of Computer Science and Artificial Intelligence, Fudan University}
  \city{Shanghai}
  \country{China}
}

\author{Xiaoqiang Li}
\email{xqli@shu.edu.cn}
\affiliation{%
  \institution{The School of Computer Engineering and Science, Shanghai University}
  \city{Shanghai}
  \country{China}
}

\author{Wei Zhang}
\email{weizh@fudan.edu.cn}
\authornote{Corresponding authors.}
\affiliation{%
  \institution{Shanghai Key Lab of Intelligent Information Processing, College of Computer Science and Artificial Intelligence, Fudan University}
  \city{Shanghai}
  \country{China}
}

\begin{abstract}
Unsupervised Video Object Segmentation (UVOS) aims to predict pixel-level masks for the most salient objects in videos without any prior annotations. While memory mechanisms have been proven critical in various video segmentation paradigms, their application in UVOS yield only marginal performance gains despite sophisticated design. Our analysis reveals a simple but fundamental flaw in existing methods: \textbf{over-reliance on memorizing high-level semantic features}. UVOS inherently suffers from the deficiency of lacking fine-grained information due to the absence of pixel-level prior knowledge. Consequently, memory design relying solely on high-level features, which predominantly capture abstract semantic cues, is insufficient to generate precise predictions. To resolve this fundamental issue, we propose a novel hierarchical memory architecture to incorporate both shallow- and high-level features for memory, which leverages the complementary benefits of pixel and semantic information. Furthermore, to balance the simultaneous utilization of the pixel and semantic memory features, we propose a heterogeneous interaction mechanism to perform pixel-semantic mutual interactions, which explicitly considers their inherent feature discrepancies. Through the design of Pixel-guided Local Alignment Module (PLAM) and Semantic-guided Global Integration Module (SGIM), we achieve delicate integration of the fine-grained details in shallow-level memory and the semantic representations in high-level memory. Our \textbf{H}ierarchical \textbf{M}emory with \textbf{H}eterogeneous \textbf{I}nteraction Network (\textbf{HMHI-Net}) consistently achieves state-of-the-art performance across all UVOS and video saliency detection benchmarks. Moreover, HMHI-Net consistently exhibits high performance across different backbones, further demonstrating its superiority and robustness. Project page: https://github.com/ZhengxyFlow/HMHI-Net . 
\end{abstract}

\keywords{Unsupervised video object segmentation, memory mechanism, optical flow, pixel-level feature, semantic feature}

\maketitle

\begin{figure}[t]
  \centering
  
  \begin{subfigure}[t]{0.95\linewidth}
    \centering
    \includegraphics[width=\linewidth]{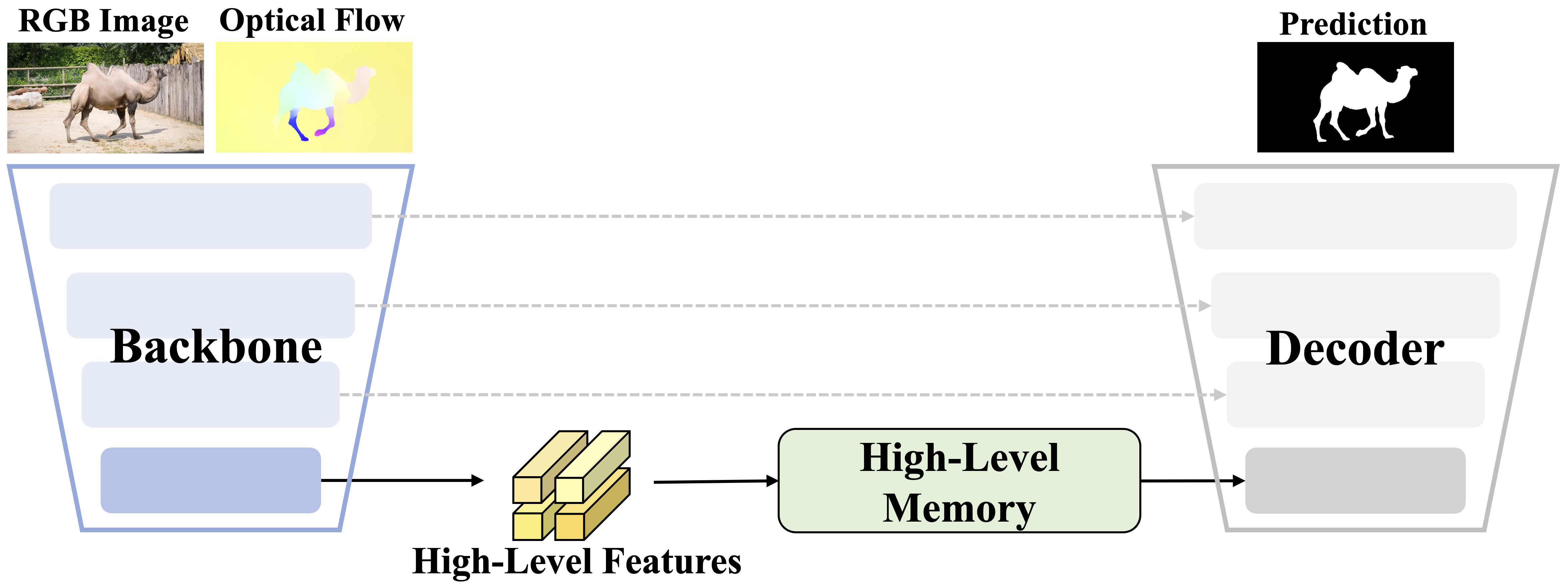}
    \caption{Single-Level Memory Architecture}
    \label{fig:single-memory}
  \end{subfigure}
  \vspace{0.5em}

  \begin{subfigure}[t]{0.95\linewidth}
    \centering
    \includegraphics[width=\linewidth]{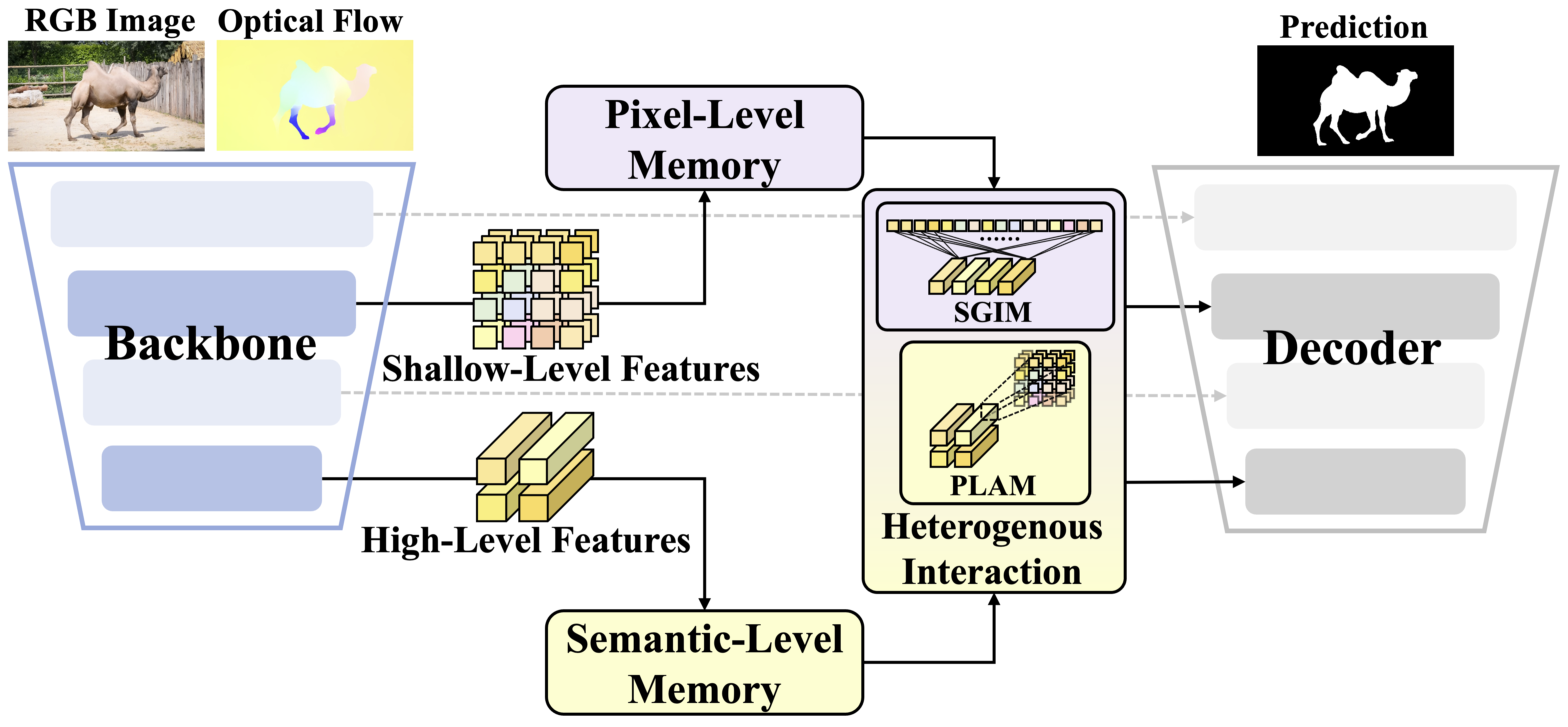}
    \caption{Hierarchical Memory Architecture}
    \label{fig:multi-memory}
  \end{subfigure}

  \caption{Illustration of the conventional single-level memory architecture and our hierarchical memory architecture. (a) Single-level memory architecture which solely relies on high-level features. (b) Hierarchical memory architecture which incorporates both shallow- and high-level features for memory.}
  \label{fig:Mem structure}
\end{figure}

\section{Introduction}

Unsupervised video object segmentation (UVOS) aims to segment the most salient object in a video sequence without any prior annotations, which makes itself a highly challenging task in visual domain. Given its ability to autonomously identify and track objects, UVOS plays a crucial role in a myriad of real-world applications.

UVOS approaches have long been confronted with a fundamental challenge: predicting pixel-wise precise segmentation with no prior knowledge. To address this issue, mainstream UVOS methods\cite{MATNet, 3DCSeg, RTNet, FSNet, TransportNet, AMC-Net, HFAN, PMN, TMO, SimulFlow, DPA, GFA} commonly incorporate optical flow as an auxiliary input to guide segmentation, and focus on designing sophisticated fusion modules to enhance performance. However, optical flow only contains short-term motion cues from two consecutive frames,  which neglects the crucial long-term correspondences in video sequences. Memory mechanisms \cite{hm1,hm2,hm3,hm4,hm5,STM,STCN,XMem,OneVOS} have emerged as a powerful design in various video segmentation tasks due to their ability to effectively capture temporal dependencies across the video sequence. Some recent approaches\cite{TGFormer,PMN, HGPU, DPA, GSA} have explored the integration of long-term memory mechanisms into UVOS. Nevertheless, these memory-based methods have yielded only marginal performance gains despite their intricate architectures. We observe a simple yet pivotal defect in these methods: \textbf{A predominant reliance on high-level memory features, accompanied by a disregard for the fundamental limitations intrinsic to UVOS.}

\begin{figure}[!t]
    \centering
    \setlength{\tabcolsep}{1pt}
    \begin{tabular}{ccccc}
        &\textbf{Image} & \textbf{Mask} & \textbf{Optical Flow} & \textbf{Attention} \\ 
        \rotatebox{90}{\hspace{0.15cm} \textbf{Level 1}} & 
        \includegraphics[width=0.23\linewidth, height=1.3cm]{Image/Section1/attention/level1/img.png} & 
        \includegraphics[width=0.23\linewidth, height=1.3cm]{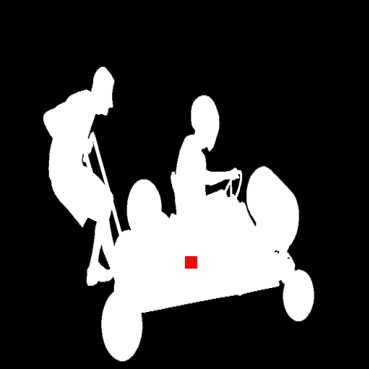} & 
        \includegraphics[width=0.23\linewidth, height=1.3cm]{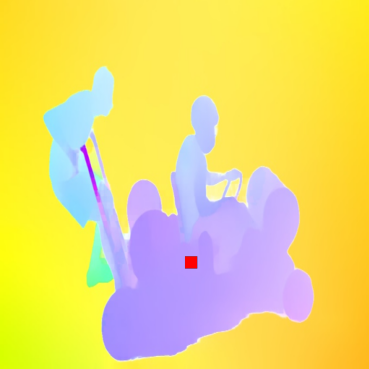} &
        \includegraphics[width=0.23\linewidth, height=1.3cm]{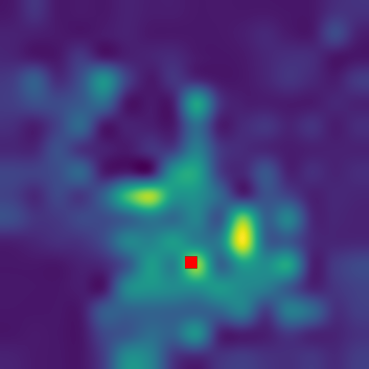} \\ 
        \rotatebox{90}{\hspace{0.15cm} \textbf{Level 2}} & 
        \includegraphics[width=0.23\linewidth, height=1.3cm]{Image/Section1/attention/level2/img.png} & 
        \includegraphics[width=0.23\linewidth, height=1.3cm]{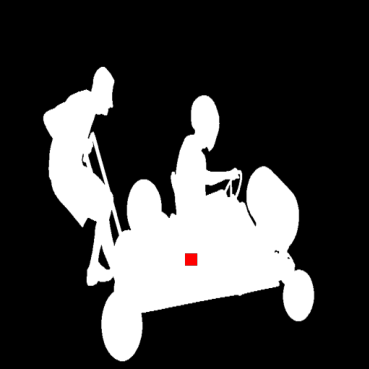} & 
        \includegraphics[width=0.23\linewidth, height=1.3cm]{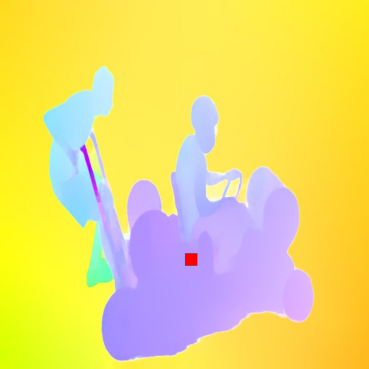} &
        \includegraphics[width=0.23\linewidth, height=1.3cm]{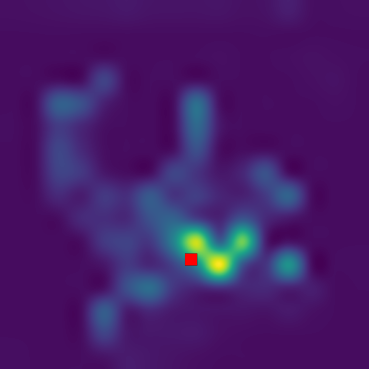} \\ 
        \rotatebox{90}{\hspace{0.15cm} \textbf{Level 3}} & 
        \includegraphics[width=0.23\linewidth, height=1.3cm]{Image/Section1/attention/level3/img.png} & 
        \includegraphics[width=0.23\linewidth, height=1.3cm]{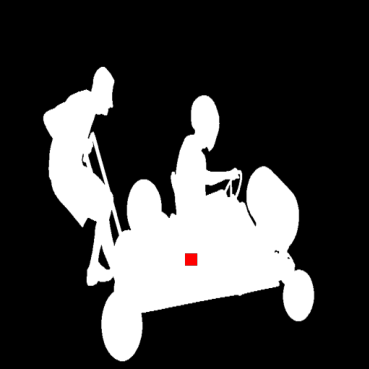} & 
        \includegraphics[width=0.23\linewidth, height=1.3cm]{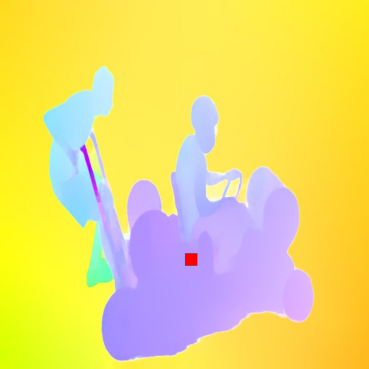} &
        \includegraphics[width=0.23\linewidth, height=1.3cm]{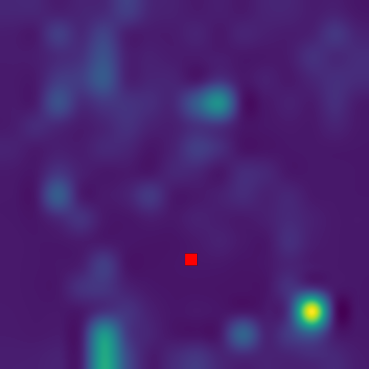} \\ 
        \rotatebox{90}{\hspace{0.15cm} \textbf{Level 4}} & 
        \includegraphics[width=0.23\linewidth, height=1.3cm]{Image/Section1/attention/level4/img.png} & 
        \includegraphics[width=0.23\linewidth, height=1.3cm]{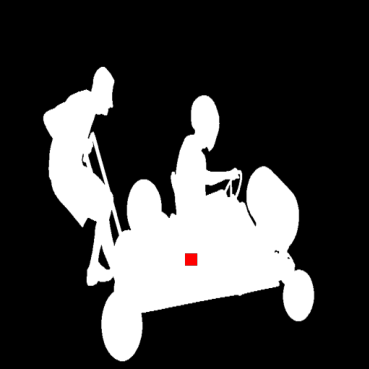} & 
        \includegraphics[width=0.23\linewidth, height=1.3cm]{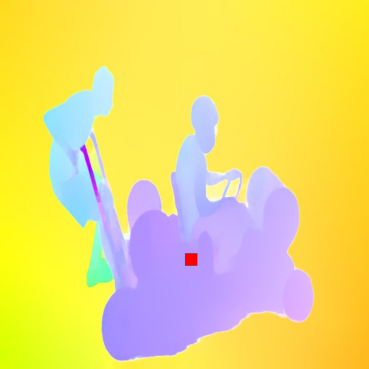} &
        \includegraphics[width=0.23\linewidth, height=1.3cm]{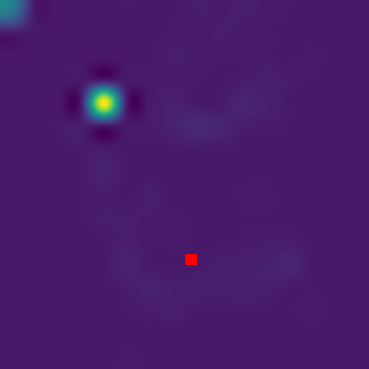} \\ 
    \end{tabular}
    \caption{Visualization of attention maps at different encoder levels. From level 1 to level 4, the model gradually focuses more on few tokens with representative semantic cues, while emphasizes less on the object details. The little read square on each map denotes the token selected for attention analysis.}
    \label{fig:Different level focus}
\end{figure}

Unlike semi-supervised video object segmentation (SVOS), where a pixel-wise mask of the first frame is provided as guidance, UVOS inherently lacks fine-grained object details and thereby struggles to generate pixel-wise predictions. Moreover, the compressing of raw images into compact high-level features at the multi-scale encoder further aggravates the loss of fine-grained details. And information retrieved from the high-level memory bank is gradually diluted during the bottom-up decoding phase. We investigate the information focus of different layers during the encoding process, and visualize their attention maps in Fig. \ref{fig:Different level focus}. It can be observed that shallow encoding levels (level 1 and 2) focus more on the general pixels of foreground objects, while high encoding levels (level 3 and 4) only emphasize on few key points which best conveys object semantics. As a result, high-level memory alone can hardly compensate for the intrinsic absence of pixel-level guidance in UVOS, leading to segmentation maps with imprecision and insufficient details. This shortcoming remarkably limits the performance of previous UVOS models, particularly in complex scenarios.

To tackle the aforementioned challenge, we put forwardthe Hierarchical Memory with Heterogeneous Interaction Network (HMHI-Net) for UVOS. Firstly, we propose a simple yet effective hierarchical memory architecture which innovatively integrates both high-level and shallow-level features for memory. High-level features, which primarily encode semantic information, contribute to maintaining object consistency across frames. In contrast, shallow-level features preserve rich pixel-wise details, thereby enhancing the segmentation of fine-grained structures. We construct two separate memory banks for both features, and thereby optimize feature encoding with both semantic and pixel-level from memorized frames. Additionally, predicted masks are added to the memory banks during the memory update. By leveraging both semantic and pixel-level memory, the proposed architecture realizes frame-wise feature refinement through hierarchical guidance and produces segmentation maps with remarkable advancements. 

Furthermore, we introduce a heterogeneous interaction mechanism to balance the contributions of the two memory banks and facilitate their feature mutual refinement. Due to the feature discrepancy between the shallow- and high-level memory, the improper use of their features can lead to feature misalignment and performance degradation. We address this issue by conducting shallow-high mutual interactions to facilitate their bidirectional feature refinement, which at the same time explicitly accounts for their inherent feature distinctions. Specifically, shallow-level features emphasize more on local fine-grained details, while high-level features capture global semantic representations. Accordingly, we design two specialized modules: the Pixel-guided Local Alignment Module (PLAM) and the Semantic-guided Global Integration Module (SGIM), tailored for shallow-to-high and high-to-shallow refinement, respectively. PLAM performs shallow-to-high information integration according to the relative positions of tokens, thereby preserving spatial coherence of pixel-level details and minimizing interference from unrelated details. Conversely, SGIM realizes high-to-shallow advancement by enabling broader perception filed on semantic cues in high-level tokens, which better leverages the comprehensive semantic guidance. With PLAM and SGIM, the heterogeneous interaction mechanism achieves a well-balanced shallow-high mutual interactions, and effectively optimizes both features with their complementary nature, remarkably elevating the overall model performance.

Our contributions can be summarized as follows:

• We propose a novel hierarchical memory architecture that simultaneously incorporates shallow- and high-level features for memory, facilitating UVOS with both pixel-level details and semantic richness stored in memory banks.

• We introduce the pixel-guided local alignment module (PLAM) and the semantic-guided global integration module (SGIM), which perform heterogeneous mutual refinement between high-level and low-level features according to their feature distinctions.

• Our HMHI-Net achieves state-of-the-art performance on all UVOS and video salient object detection (VSOD) benchmarks, with 89.8\% \(\mathcal{J} \& \mathcal{F}\) on DAVIS-16\cite{DAVIS-16}, 86.9\% \(\mathcal{J}\) on FBMS\cite{FBMS} and 76.2\% \(\mathcal{J}\) on YouTube-Objetcs\cite{YouTube-Objects}. Moreover, HMHI-Net consistently delivers high performance across different backbones, underscoring its superior generalization capability and robustness.

\section{Related Work}
\subsection{Semi-supervised Video Object Segmentation}
 Semi-supervised Video Object Segmentation (SVOS) aims to segment the target objects throughout the subsequent video sequence by leveraging the given object masks in the first frame. STM\cite{STM} and STCN\cite{STCN} pioneered a space-time memory that computes similarity between current and past frames to propagate masks, maintaining spatio-temporal consistency. To mitigate the increasing computational cost in long video sequences, methods such as AOT\cite{AOT}, XMem\cite{XMem}, Cutie\cite{Cutie} and SAM2\cite{SAM2} introduced a layered memory design, proposed long- and short-term memory or object tokens to compresses semantically rich features from distant frames. Other approaches, such as OneVOS\cite{OneVOS}, discarded explicit memory selection and instead input all previous frames, allowing the model to dynamically select and store informative key frame features.

 Although matching-based techniques demonstrate strong performance in SVOS, their direct application to UVOS proves challenging. In the UVOS setting, where the first-frame mask is unavailable, representations encoded from high-level features inherently lack sufficient spatial details. 

\subsection{Unsupervised Video Object Segmentation}
Unlike SVOS, Unsupervised Video Object Segmentation (UVOS) requires segmenting the most salient object in a video without prior supervision or manual annotations. Early UVOS methods \cite{PDB, MOTAdapt, AGS, COSNet, AD-Net, AGNN} primarily relied on exploiting appearance consistency across frames. 
A recent major trend in UVOS research involves leveraging optical flow to capture object motion and promotes segmentation. Representative works\cite{MATNet, 
WCS-Net,DFNet,RTNet, FSNet, TransportNet, AMC-Net, HCPN} have developed various fusion modules to integrate optical flow with image appearance. Typically, these methods adopt either two-stream\cite{TMO} or single-stream\cite{HFAN,SimulFlow} backbones to extract flow and image features, which are then combined through complex fusion mechanisms. Despite their contributions, flow-based methods remain constrained by short-term motion cues, making them prone to errors under occlusion or rapid motion.

Recent works\cite{GSA, DPA, TGFormer, PMN} attempt to address these limitations by incorporating long-term memory in UVOS. These methods typically fuse high-level visual and motion features from reference frames with the current frame to enhance segmentation. However, they only utilize the high-level features of the reference frames, failing to capture more fine-grained pixel-level information. To address this, we store both shallow- and high-level features in memory, supplying both precise details and strong semantic priors for accurate, consistent segmentation.

\section{Methodology}

\begin{figure*}[!t]
\centering
\includegraphics[width=7in,height=4.8in]{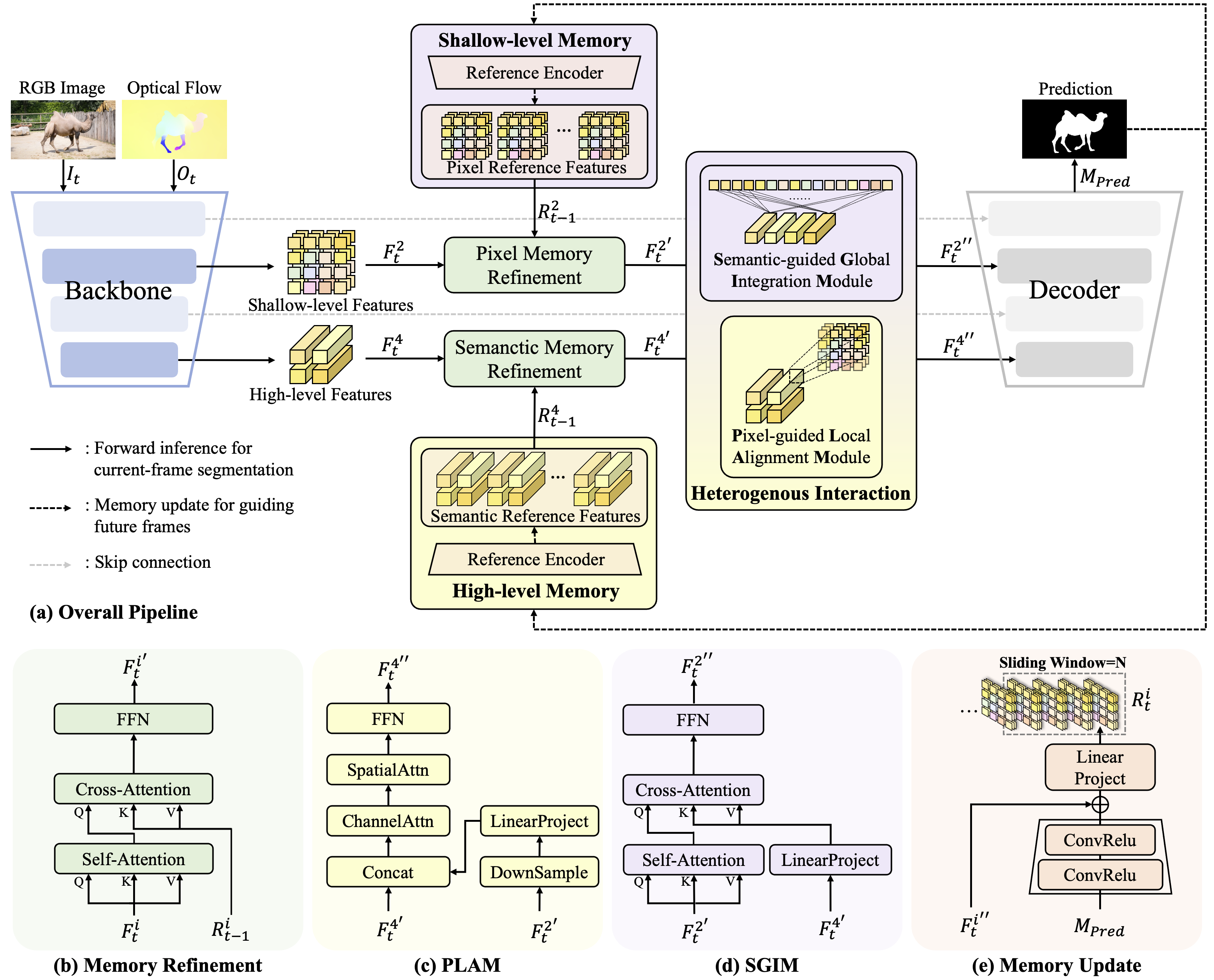}
\caption{(a) Overall pipeline of HMHI-Net. (b) Memory readout mechanism to refine current frame. (c) Pixel-guided local alignment module. (d)Semantic-guided global integration module. (e) Memory update mechanism with the reference encoder. }
\label{fig:Overal_Pipeline}
\end{figure*}

\subsection{Overall Pipeline}
We employ a hierarchical backbone as the encoder following the conventional segmentation paradigm, which generates multi-scale features for decoding. At time $t$, the encoder takes the current frame image $I_t \in \mathbb{R}^{H \times W \times 3}$ and its corresponding optical flow map $O_t \in \mathbb{R}^{H \times W \times 3}$ as inputs, and extracts multi-scale features through four hierarchical layers. $I_t^i \in \mathbb{R}^{H_i W_i \times C_i}$ and $O_t^i \in \mathbb{R}^{H_i W_i \times C_i}$ ($i \in \{1, 2, 3, 4\}$) represents the encoded feature of image and optical flow at $i$ -th layer, where \( H_iW_i = \frac{H}{2^{i+1}} \times \frac{W}{2^{i+1}} \). We simply add $I^t_i$ and $O^t_i$ from the same level to form the merged feature $F_t^i \in \mathbb{R}^{H_i W_i \times C_i}$ for further processing. The general encoding phase is briefly presented as follows:

\begin{equation}
\label{general encoder}
\begin{aligned}
    I_t^i,\, O_t^i &= \text{Encoder}(I_t, O_t) \\
    F_t^i &= I_t^i + O_t^i, \quad i \in \{1, 2, 3, 4\}
\end{aligned}
\end{equation}

During the encoding phase, the encoded feature $F_t^2$ and $F_t^4$ are further optimized by extracting helpful knowledge from their relevant memory banks. To be more specific, the memory bank stores numerous history frames as the reference features $R^i_{t-1} \in \mathbb{R}^{T H_i W_i \times C_i}$ ($i\in\{2,4\}$), where $T$ is the number of memorized frames and $t-1$ is the time step denoting the reference for the current frame at time $t$. As presented in \eqref{simple show of memory readout}, we adopt a unified refinement architecture for both levels to attend to either spatial details or semantic object cues from their individual reference features.

\begin{equation}
\label{simple show of memory readout}
\begin{aligned}
F_t^{2'} &= \text{Mem\_Refine}(F_t^2, R_{t-1}^2)\\
F_t^{4'} &= \text{Mem\_Refine}(F_t^4, R_{t-1}^4)
\end{aligned}
\end{equation}

We further propose the heterogeneous interaction mechanism to promote mutual refinement of $F_t^{2'}$ and $F_t^{4'}$. Specifically, we propose the pixel-guided local alignment module (PLAM) for shallow-to-high refinement. PLAM adopts structure-preserving attention mechanism to retrieve fine-grained knowledge, which preserves the spatial layout and fine-grained structural information from the shallow-level feature $F_t^{2'}$. Additionally, the semantic-guided global integration module (SGIM) applies a global attention strategy to extract semantic cues from the high-level feature map $F_t^{4'}$ and aligns them with the shallow-level representation $F_t^{2'}$. The mutual refinement process is formulated as:

\vspace{-2mm}
\begin{equation}
\label{PRLF and SGGF}
\begin{aligned}
F_t^{2''} &= \text{SGIM}(F_t^{4'}, F_t^{2'}) \\
F_t^{4''} &= \text{PLAM}(F_t^{2'}, F_t^{4'})
\end{aligned}
\end{equation}
\vspace{-2mm}

We then use the updated multi-scale features $[F_t^1, F_t^{2''}, F_t^3, F_t^{4''}]$ as inputs to the hierarchical decoder. The decoder progressively upsamples the features from the top layer, fusing them with lower-level features in a bottom-up manner and ultimately generates the segmentation map $M_{\text{Pred}} \in \mathbb{R}^{H \times W \times 1}$. And $M_{\text{Pred}}$ is added to the memory banks together with the encoded features for memory update as follows:

\begin{equation}
\label{decode and memorize}
\begin{aligned}
& M_{\text{Pred}} = \text{Decoder}(F_t^1, F_t^{2''}, F_t^3, F_t^{4''})\\
& R_{t}^2 = \text{Mem\_Update}(R_{t-1}^2, F_t^{2''}, M_{\text{Pred}})\\
& R_{t}^4 = \text{Mem\_Update}(R_{t-1}^4, F_t^{4''}, M_{\text{Pred}})\\
\end{aligned}
\end{equation}

\subsection{Hierarchical Memory Structure}
We construct our hierarchical memory architecture based on the four-layer hierarchical encoder. Normally, the first two layers are considered as the shallow level, which concentrate more on local information. And the last two layers are deemed as the high level, which mostly encode abstract representations. As illustrated in Fig.\ref{fig:Different level focus}, the second-layer feature $F_t^2$ possesses abundant pixel-level information and is encoded more sufficiently compared to $F_t^1$, which contains misleading focus on background pixels. Conversely, the fourth-layer features $F_t^4$ conveys the most compact semantic cues. Therefore, we choose the second and fourth layers to establish two separate memory banks, which correspond to the shallow- and high-level memory, respectively. 

Given a current frame image $I_t$ and its optical flow $O_t$, we optimize the encoded features $F_t^i$ ($i \in \{2, 4\}$) using their corresponding reference features $R^i_{t} \in \mathbb{R}^{T H_i W_i \times C_i}_{t-1}$ ($i\in\{2,4\}$) from memory. In order to  avoid misalignment between shallow- and high-level features, we apply the same memory mechanism to these separate memory banks. First of all, we first employ attention mechanism to feature $F_t^i$ ($i \in \{2, 4\}$) itself, to enhance $F_t^i$ by attending to its own internal representations. As expressed in \eqref{memory self-atten}, we projected $F_t^i$ into query $Q_{\text{sa}}^{i,t}$, key $K_{\text{sa}}^{i,t}$, and value $V_{\text{sa}}^{i,t}$ embeddings via three separate linear layers, and perform scaled dot-product attention accordingly.

\begin{equation}
\label{memory self-atten}
\begin{aligned}
F_t^{i'} & = \text{Attention}(Q_{\text{sa}}^{i,t}, \, K_{\text{sa}}^{i,t}, \, V_{\text{sa}}^{i,t}) \\[0.6em]
& = \text{Softmax}\left(\frac{Q_{\text{sa}}^{i,t \top} K_{\text{sa}}^{i,t}}{\sqrt{d}}\right)V_{\text{sa}}^{i,t}
\end{aligned}
\end{equation}

We further use the stored reference characteristics $R^i_{t-1}$ at time $t$ to optimize current feature $F_t^{i'}$. Memory banks contains the encode attributes of former predictions, which are relatively reliable supervision. We derive $Q_{\text{mem}}^{i,t}$ from the present feature $F_t^{i'}$, and $K_{\text{mem}}^{i,t}$, $V_{\text{mem}}^{i,t}$ from $R^i_{t}$, and calculate the per-token relationship between $F_t^{i'}$ and $R^i_{t}$ as shown in \eqref{memory cross-attn}.

\begin{equation}
\label{memory cross-attn}
\begin{aligned}
S_{\text{corr}} &= (Q_{\text{mem}}^{i,t \top} \: K_{\text{mem}}^{i,t} )\: /  \: \sqrt{d}
\end{aligned}
\end{equation}

The attention map $S_{\text{corr}} \in \mathbb{R}^{H_2 W_2 \times T H_2 W_2} $ reveals the relation between the current frame and the reference frames, where each row indicates the similarity between a pixel in the current frame and all pixels in $T$ reference frames. We normalize $S_{\text{corr}}$ through a softmax function and retrieve information accordingly from $V_{\text{mem}}^{i,t}$ to refine $F_t^{i'}$. The read out attributes are added back to $F_t^{i'}$. Additionally, a feed-forward neural network (FFN) is applied to $F_t^{i'}$ to realign the feature vector back to the same representational space as those produced by the backbone. This memory readout process is formulated in \eqref{memory readout}.

\begin{equation}
\label{memory readout}
\begin{aligned}
F_t^{i'} + & = \text{Softmax}\left(S_{\text{corr}}\right) V_{\text{mem}}^{i,t}\\
F_t^{i'} & = \text{FFN}(F_t^{i'})
\end{aligned}
\end{equation}

\subsection{Heterogeneous Interaction Mechanism}
To avoid feature misalignment between the shallow- and high-level memory banks due to their feature discrepancy, we propose the heterogeneous interaction mechanism to facilitate the pixel-semantic memory mutual interactions. The heterogeneous interaction mechanism takes advantage of the complementary nature of $F_t^{2'}$ and $F_t^{4'}$, and designs different modules for the shallow-to-high and high-to-shallow feature communications according to their character differences. Specifically, shallow features focus more on local patches, and relative positional relation is important when retrieving information from it. High-level features have more inclusive representations, which requires a broader perception field to fully exploit the semantic cues. The heterogeneous interaction mechanism consists of two key components: the pixel-guided local alignment module (PLAM) and the semantic-guided global integration module (SGIM).

\textbf{Pixel-guided Local Alignment Module}. PLAM is introduced to enrich the high-level feature $F_t^{4'} \in \mathbb{R}^{H_4W_4 \times C_4}$ with detailed structural information extracted from the shallow-level feature $F_t^{2'} \in \mathbb{R}^{H_2W_2 \times C_2}$. By incorporating fine-grained local cues into the high-level representation, the decoder is guided with structural priors from the very beginning, reducing confusion caused by background regions with similar semantics. PLAM first projects $F_t^{2'}$ to align the high-level feature $F_t^{4'}$ in dimension space via a series of downsample operations, producing the aligned features $F_t^{2\_\text{tmp}} \in \mathbb{R}^{H_4W_4 \times C_4}$ as in \eqref{F2 downsample}.

\begin{equation}
\label{F2 downsample}
\begin{aligned}
F_t^{2\_\text{tmp}} & = \text{ConvReLu}\left(F_t^{2'} \right)\\
F_t^{2\_\text{tmp}} & = \text{Linear}(F_t^{2\_\text{tmp}})
\end{aligned}
\end{equation}

To preserve the spatial consistency from the shallow-level features, we directly concatenate the aligned features $F_t^{2\_\text{tmp}}$ with $F_t^{4'}$ at the $C$ dimension for unified representation $F_t^{4''} \in \mathbb{R}^{H_4W_4 \times 2C_4}$. Next, we employ the channel attention to re-emphasize channel-wise information which enhances the feature's semantic expressiveness at the channel level. This process is denoted as:

\begin{equation}
\label{channel attention}
\begin{aligned}
F_t^{4''} &= \text{Concat}(F_t^{4'}, F_t^{2\_\text{tmp}})\\
F_t^{4''} &= \text{Channel\_Attn}(F_t^{4''})
\end{aligned}
\end{equation}

Following this, spatial attention is employed to identify spatial locations that are crucial for object representation, which promotes the semantic focus on the target regions, yielding the shallow-enhanced high-level feature $F_t^{4''}$. Finally, an FFN is utilized to project the refined feature back to the original feature space as in \eqref{spatial attention}, ensuring compatibility during the decoding phase.

\begin{equation}
\label{spatial attention}
\begin{aligned}
F_t^{4''} &= \text{Spatial\_Attn}(F_t^{4''})\\
F_t^{4''} &= \text{FFN}(F_t^{4''})
\end{aligned}
\end{equation}

\textbf{Semantic-guided Global Integration Module}. To better inject high-level semantics into shallow-level features $F_t^{2'}$ and prevent dilution of semantic cues during decoding, we design the SGIM. Similarly, SGIM begins by aligning abstract high-level features $F_t^{4'} \in \mathbb{R}^{H_4W_4 \times C_4}$ to the shallow-level pixel feature space via a linear projector, producing aligned features $F_t^{4\_\text{tmp}} \in \mathbb{R}^{H_4W_4 \times C_4}$. Next, SGIM extracts \(Q_{\text{sa}}^{2,t}, K_{\text{sa}}^{2,t}, V_{\text{sa}}^{2,t}\) from $F_t^{2'}$ and applies the attention mechanism to refine $F_t^{2'}$ by modeling stronger inner pixel-level relations, as expressed in \eqref{F2 self-atten}.

\begin{equation}
\label{F2 self-atten}
\begin{aligned}
& F_t^{4\_\text{tmp}}  = \text{Linear}(F_t^{4'}) \\[0.6em]
& F_t^{2'}  = \text{Attention}(Q_{\text{sa}}^{2,t}, \, K_{\text{sa}}^{2,t}, \, V_{\text{sa}}^{2,t}) \\[0.6em]
& \quad \;= \text{Softmax}\left(\frac{Q_{\text{sa}}^{2,t \top}  K_{\text{sa}}^{2,t}}{\sqrt{d}}\right)V_{\text{sa}}^{2,t}
\end{aligned}
\end{equation}

Although the aligned high-level features $F_t^{4\_\text{tmp}}$ share the same dimensionality as $F_t^{2'}$, they differ in spatial resolution and semantic abstraction. To integrate global semantic context into each pixel representation, we apply a global attention mechanism to $F_t^{4\_\text{tmp}}$ and $F_t^{2'}$, where $Q_{\text{ca}}^{2,t}$ is derived from $F_t^{2'}$ and $K_{\text{ca}}^{4,t}$, $V_{\text{ca}}^{4,t}$ are projected from $F_t^{4\_\text{tmp}}$:

\begin{equation}
\label{SGIM cross attention}
\begin{aligned}
F_2'' +&= \text{Attention}(Q_{\text{ca}}^{2,t}, K_{\text{ca}}^{4,t},V_{\text{ca}}^{4,t})\\[0.3em]
&= \text{Softmax}\left(\frac{Q_{\text{ca}}^{2,t}\,  K_{\text{ca}}^{4,t}}{\sqrt{d}}\right)V_{\text{ca}}^{4,t}
\end{aligned}
\end{equation}

Finally, an FFN is applied to map the refined shallow-level features $F_2''$ back to their original representation space, completing the fusion process from object-level semantics to pixel-level details.

\subsection{Memory Update}
After generating the predicted mask $M_{\text{Pred}}$ of the current frame, we update two memory banks the final refined features $F_2''$ and $F_4''$, together with the predicted mask. Two simple memory encoders are employed to integrate $M_{\text{Pred}}$ into the shallow- and high-level refined features. 

During memory bank updates, we adopt a sliding window strategy with maximum memory limit $N$. The memory bank $R^i_{t} \in \mathbb{R}^{T H_i W_i \times C_i}$, where $T \in \{1, 2, \ldots, N\}$, stores the most recent $T$ reference features. We update $R^i_{t}$ every $k$ frames following the first-in-first-out manner. Since no reference frame is available for the first frame in a video sequence, we only utilize the simple baseline without any memory refinement or heterogeneous interaction.

\section{Experiments}
\subsection{Implementation Details}
\textbf{Training and Inference}. Following \cite{HFAN,SimulFlow,DTTT,ISTC-Net}, we utilize mit\_b1 as our backbone for fair comparison. We adopt the simple and efficient motion-appearance integration paradigm in \cite{ISTC-Net} to avoid redundant discussion on the fusion mechanism, which is not an emphasis in this paper. We directly employ the common multi-scale decoder as \cite{TMO,HFAN} for fair comparison. At the training stage, a sequence of five frames is selected for training in each iteration, where $k = 1$ and $T=5$. The first frame skips the memory refinement and heterogeneous interaction modules, and only stores the shallow- and high-level features into their respective memory banks. We follow \cite{ISTC-Net, SAM2} to adopt a combination of binary cross entropy loss, focal loss and dice loss for training. The final training loss is computed as the average of the segmentation losses from all five frames.

In the training stage, we adopt the AdamW optimizer with a learning rate of 6e-5, and train the model for 150 epochs on the YouTube-VOS datasets\cite{YouTube-VOS}. During fine-tuning, we set the learning rate to 1e-4 with a CosineAnnealingLR scheduler and train the model until convergence. All training and inference are conducted on four NVIDIA RTX 4090 GPUs.

For inference, $k$ is set to 1 and $T$ is set to 5 across all benchmarks for convenience as in prior works. Images are resized to $512 \times 512 $ during both training and inference.

\textbf{Evaluation Metrics}. We assess model performance using a comprehensive set of metrics. For UVOS, we adopt region similarity $\mathcal{J}$, which evaluates segmentation accuracy via intersection-over-union (IoU). Boundary accuracy $\mathcal{F}$ measures the quality of mask contours through F1 score computation. Their average, \(\mathcal{J} \& \mathcal{F}\), serves as the overall performance indicator. 

For VSOD, we report mean absolute error (MAE) to quantify pixel-level prediction accuracy, maximum F-measure ($F_m$) to capture the best precision-recall tradeoff, enhanced-alignment measure ($E_m$) to reflect both pixel-wise and global consistency, and structure-measure ($S_m$) to evaluate region-aware and object-aware structural similarity.

\begin{table*}[!ht]
    \centering
    \caption{Evaluation results on three UVOS benchmarks: DAVIS-16, FBMS, and YouTube-Objects. Methods employing optical flow are marked with 'OF', while 'PP' indicates the use of post-processing. The top-performing and runner-up methods are emphasized using \textbf{bold} and \underline{underline} formatting, respectively.}
    \renewcommand{\arraystretch}{0.8}
    \begin{tabularx}{\textwidth}{p{2.5cm}p{2cm}p{2.5cm}XXXXXXX} 
        \hline
        \multirow{2}{*}{\textbf{Method}} & \multirow{2}{*}{\textbf{Publication}} & \multirow{2}{*}{\textbf{Backbone}} & \multirow{2}{*}{\textbf{OF}} & \multirow{2}{*}{\textbf{PP}} & \multicolumn{3}{c}{\textbf{DAVIS-16}} & \textbf{FBMS} & \textbf{YTO} \\ 
        \cline{6-10}
        & & & & & \(\mathcal{J} \& \mathcal{F}\) & \(\mathcal{J}\) & \(\mathcal{F}\) & \(\mathcal{J}\) & \(\mathcal{J}\)\\
        \hline
        PDB\cite{PDB} & ECCV'18 & ResNet-50 & &  \checkmark &75.9 & 77.2 & 74.5 & 74.0 & -  \\ 
        COSNet\cite{COSNet} & CVPR'19 & DeepLabv3 & & \checkmark & 80.0 & 80.5 & 79.4 & 75.6 & 70.5 \\ 
        AGNN\cite{AGNN} & ICCV'19 & DeepLabV3 & & \checkmark  & 79.9 & 80.7 & 79.1 & - & 70.8 \\ 
        MATNet\cite{MATNet} & AAAI'20 & ResNet-101 & \checkmark & \checkmark & 81.6 & 82.4 & 80.7 & 76.1 & 69.0 \\ 
        DFNet\cite{DFNet} & ECCV'20 & DeepLabv3 & & \checkmark &  82.6 & 83.4 & 81.8 & - & - \\ 
        RTNet\cite{RTNet} & CVPR'21 & ResNet-101 & \checkmark &\checkmark & 85.2 & 85.6 & 84.7 & - & 71.0 \\ 
        TransportNet\cite{TransportNet} & ICCV'21 & ResNet-101 & \checkmark & & 84.8 & 84.5 & 85.0 & 78.7 & - \\ 
        AMC-Net\cite{AMC-Net} & ICCV'21 & ResNet-101 & \checkmark & \checkmark & 84.6 & 84.5 & 84.6 &  76.5 & 71.1 \\ 
        IMP\cite{IMP} & AAAI'22 & ResNet-50 &  &  & 85.6 & 84.5 & 86.7 & 77.5 & -\\ 
        HFAN\cite{HFAN} & ECCV'22 & Mit-b1 & \checkmark &  & 86.7 & 86.2 & 87.1 & - & 73.4 \\ 
        HCPN\cite{HCPN} & TIP'23 & ResNet-101 & \checkmark & \checkmark & 85.6 & 85.8 & 85.4 & 78.3 & 73.3 \\
        PMN\cite{PMN} & WACV'23 & VGG-16 & \checkmark &  & 85.9 & 85.4 & 86.4 & 77.7 & 71.8 \\ 
        TMO\cite{TMO} & WACV'23 & ResNet-101 & \checkmark &  & 86.1 & 85.6 & 86.6 & 79.9 & 71.5\\ 
        OAST\cite{OAST} & ICCV'23 & MobileViT & \checkmark & & 87.0 & 86.6 & 87.4 & 83.0 & - \\
        TGFormer\cite{TGFormer} & ACMMM'23 & MobileViT & & & 86.3 & 85.8 & 86.7 & 84.0 & -  \\
        SimulFlow\cite{SimulFlow} & ACMMM'23 & Mit-b1 & \checkmark &  & 87.4 & 86.9 & 88.0 & 80.4 & 72.9 \\ 
        HGPU\cite{HGPU} & TIP'24 &  ResNet-101 & \checkmark &  & 86.1 & 86.0 & 86.2 & - & 73.9 \\
        DPA\cite{DPA} & CVPR'24 & VGG-16 & \checkmark &  & 87.6 & 86.8 & 88.4 & \underline{83.4} & 73.7\\  
        GSA\cite{GSA} & CVPR'24 & ResNet-101 & \checkmark &  & 87.7 & 87.0 & 88.4 & 83.1 & -\\ 
        DTTT\cite{DTTT} & CVPR'24 & Mit-b1 & \checkmark &  & 87.2 & 85.8 & 88.5 & 78.8 & -\\  
        GFA\cite{GFA} & AAAI'24 & - & \checkmark &  & \underline{88.2} & \underline{87.4} & \underline{88.9} & 82.4 & \underline{74.7} \\ 
        GFA\cite{GFA} & AAAI'24 & ResNet-101 & \checkmark &  & 86.3 & 85.9 & 86.7 & 80.1 & 73.6 \\
        \hline
        \textbf{Ours} & - & Mit-b1 & \checkmark &  & \textbf{89.8} & \textbf{88.6} & \textbf{91.0} & \textbf{86.9} & \textbf{76.2} \\ 
        \hline
    \end{tabularx}
    \label{tab:UVOS_method_comparison}
\end{table*}

\subsection{Quantitative Results}
\textbf{UVOS Performance}. We first pretrain the model on YouTube-VOS\cite{YouTube-VOS} and finetune the pre-trained model on DAVIS-16\cite{DAVIS-16} or FBMS\cite{FBMS} datasets for evaluations. We directly adopts the model in the pre-training stage to test on YouTube-Object\cite{YouTube-Objects}. We compare our proposed model with previous UVOS approaches across these benchmarks. As shown in Table~\ref{tab:UVOS_method_comparison}, our model achieves state-of-the-art performance on three widely used UVOS benchmarks: DAVIS-16\cite{DAVIS-16}, FBMS\cite{FBMS}, and YouTube-Objects\cite{YouTube-Objects}, and outperforms the most recent state-of-the-art methods by 1.6\%, 3.5\%, and 1.5\% respectively. By analyzing Table~\ref{tab:UVOS_method_comparison}, we observe that among all previous memory-free approaches, algorithms like FSNet\cite{FSNet} and TransportNet\cite{TransportNet} which rely solely on high-level motion-appearance fusion typically underperform compared to those methods incorporating multi-level feature alignment, such as HFAN\cite{HFAN}, TMO\cite{TMO}, and SimulFlow\cite{SimulFlow}. This implicitly highlighted the inadequacy of high-level features along for generating satisfactory results. Furthermore, memory-based methods, such as \cite{PMN, TGFormer, DPA}, achieve some advancements, which highlights the contribution of long-term memory in providing richer information. Our HMHI-Net incorporates both shallow- and high-level features into the long-tern memory, which enables robust and precise boundary segmentation under challenging and rapidly changing scenarios.

\begin{table*}[t]
    \centering
    \caption{Quantitative comparison on the VSOD benchmarks: DAVIS-16, DAVSOD, ViSal, and FBMS. In the table, results marked with * are reproduced using the official released code. $\uparrow$ denotes that higher values indicate better performance, while $\downarrow$ implies the opposite. Numbers indicated in \textbf{bold} and \underline{underline} represent the best and second-best scores, respectively.}
    \renewcommand{\arraystretch}{0.9}
    \begin{tabularx}{\textwidth}{p{2.2cm}|p{0.7cm}XXX|p{0.7cm}XXX|p{0.7cm}XXX|p{0.7cm}XXX} 
        \hline
        \multirow{2}{*}{\textbf{Method}} & \multicolumn{4}{c|}{\textbf{DAVSOD}} & \multicolumn{4}{c|}{\textbf{DAVIS-16}} & \multicolumn{4}{c|}{\textbf{ViSal}} & \multicolumn{4}{c}{\textbf{FBMS}} \\ 
        \cline{2-17}
    
        & $MAE\downarrow$ & $F_m\uparrow$ & $E_m\uparrow$ & $S_m\uparrow$ & $MAE\downarrow$ & $F_m\uparrow$ & $E_m\uparrow$ & $S_m\uparrow$ & $MAE\downarrow$ & $F_m\uparrow$ & $E_m\uparrow$ & $S_m\uparrow$ & $MAE\downarrow$ & $F_m\uparrow$ & $E_m\uparrow$ & $S_m\uparrow$ \\
        \hline
        MATNet*\cite{MATNet} & 0.098 & 0.628 & 0.789 & 0.707 & 0.048 & 0.752 & 0.890 & 0.776 & 0.041 & 0.891 & 0.967 & 0.863 & 0.091 & 0.751 & 0.852 & 0.760 \\
        RTNet*\cite{RTNet} & 0.068 & 0.647 & 0.782 & 0.743 & 0.012 & 0.928 & 0.978 & 0.933 & 0.019 & 0.938 & 0.975 & 0.936 & 0.057 & 0.845 & 0.892 & 0.855 \\ 
        FSNet\cite{FSNet} & 0.072 & 0.685 & 0.825 & 0.773 & 0.020 & 0.907 & 0.970 & 0.920 & - & - & - & - & 0.041 & 0.888 & 0.935 &0.890 \\
        TransportNet\cite{TransportNet} &- & - & - & - & 0.013 & 0.928 & - & -  & \textbf{0.012} & 0.953 & - & - & 0.045 & 0.885 & - & - \\
        HFAN*\cite{HFAN} & 0.078 & 0.656 & 0.795 & 0.763 & 0.014 & 0.930 & \underline{0.984} & \underline{0.939} & 0.029 & 0.860 & 0.928 & 0.891 & 0.065 & 0.794 & 0.877 & 0.818 \\
        TGFormer\cite{TGFormer} & 0.065 & 0.728 & - & 0.798 & \underline{0.011} & 0.922 & - & 0.932 & \underline{0.011} & \underline{0.955} & - & 0.952 & \underline{0.026} & \underline{0.919} & - & 0.916 \\
        HCPN*\cite{HCPN} & 0.072 & 0.684 & 0.818 & 0.774 & 0.017 & 0.923 & 0.980 & 0.932 & 0.016 & 0.942 & 0.986 & 0.945 & 0.060 & 0.850 & 0.903 & 0.851 \\
        TMO*\cite{TMO} & \underline{0.062} & \underline{0.731} & \underline{0.849} & \underline{0.805} & 0.013 & 0.925 & 0.982 & 0.936 & \underline{0.013} & 0.951 & \underline{0.989} & \underline{0.951} & 0.036 & 0.887 & \underline{0.933} & 0.893 \\
        OAST\cite{OAST} & 0.070 & 0.712 & - & 0.786 & \underline{0.011} & 0.926 & - & 0.935 & - & - & - & - & \textbf{0.025} & \underline{0.919} & - & \underline{0.917}\\
        SimulFlow\cite{SimulFlow} & 0.069 & 0.722 & - & 0.771  & \textbf{0.009} & \underline{0.936} & - & 0.937 & \textbf{0.012} & 0.943 & - & 0.946 & - & - & - & -\\
        \textbf{Ours} & \textbf{0.054} & \textbf{0.801} & \textbf{0.896} & \textbf{0.847} & \textbf{0.009} & \textbf{0.947} & \textbf{0.990} & \textbf{0.951} & \textbf{0.012} & \textbf{0.962} & \textbf{0.991} & \textbf{0.960} & 0.030 & \textbf{0.946} & \textbf{0.977} & \textbf{0.930} \\
        \hline
    \end{tabularx}
    \label{tab:VSOD_method_comparison}
\end{table*}

\textbf{VSOD Performance}. Following previous works, we further fine-tune HMHI-Net on a mixed dataset of DAVIS-16 and DAVSOD\cite{DAVSOD}, and evaluate our model on four viedo salient object detection benchmarks: DAVIS-16, FBMS, ViSal\cite{ViSal}, and DAVSOD. As shown in Table~\ref{tab:VSOD_method_comparison}, our model achieves the best performance on all datasets and evaluation metrics with great margins, except for MAE on FBMS, where it ranks the third. Models using long-term memory, such as \cite{TGFormer}, achieves better results on metrics like $MAE$ and $F_m$, indicating the strength of long-term memory in preserving global semantic consistency. Our HMHI-Net achieves top results across multiple datasets and metrics, demonstrating its outstanding capability in saliency detection.

\subsection{Qualitative Results}
\begin{figure}[!h]
    \centering
    \setlength{\tabcolsep}{1pt}
    \footnotesize
    \begin{tabular}{ccccccc}
    \rotatebox{90}{\textbf{Case1}} &
        \includegraphics[width=0.15\linewidth]{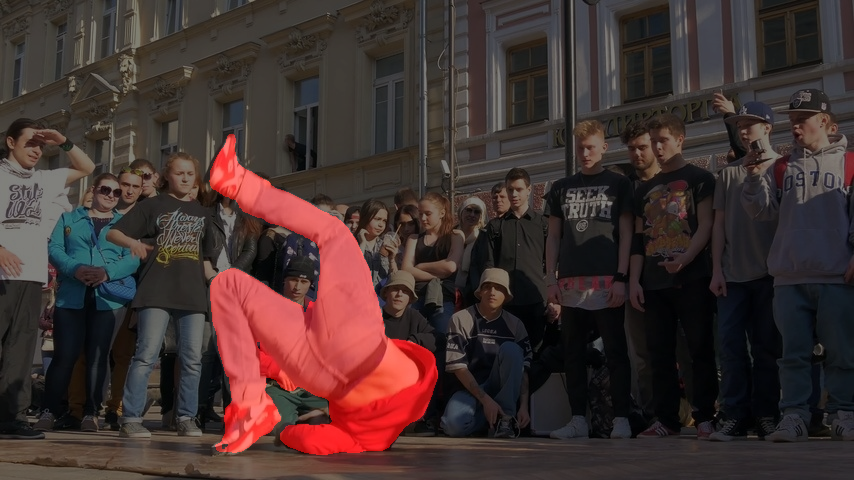} & 
        \includegraphics[width=0.15\linewidth]{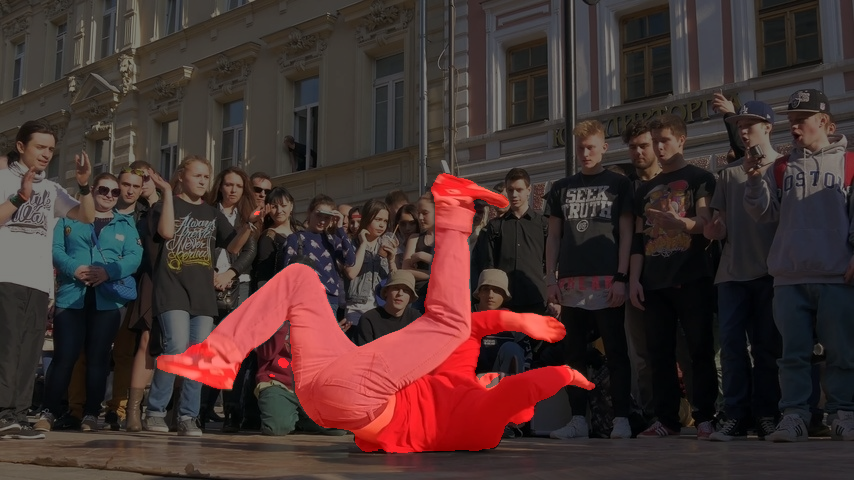} & 
        \includegraphics[width=0.15\linewidth]{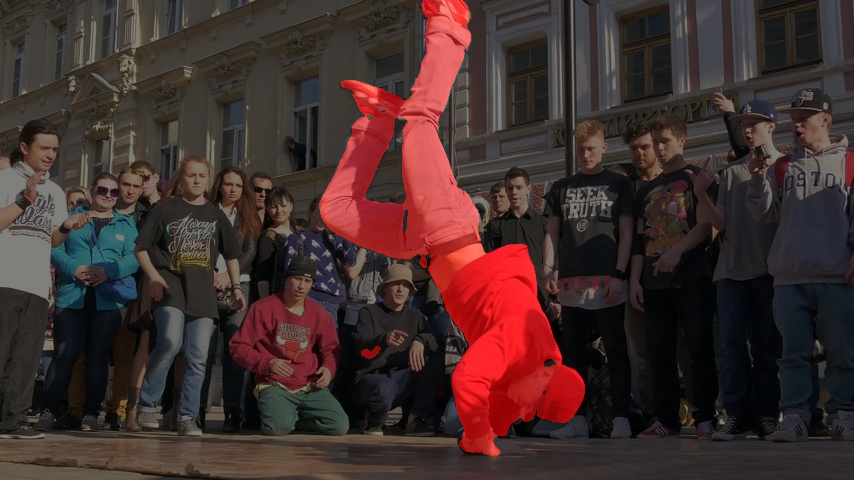} &
        \includegraphics[width=0.15\linewidth]{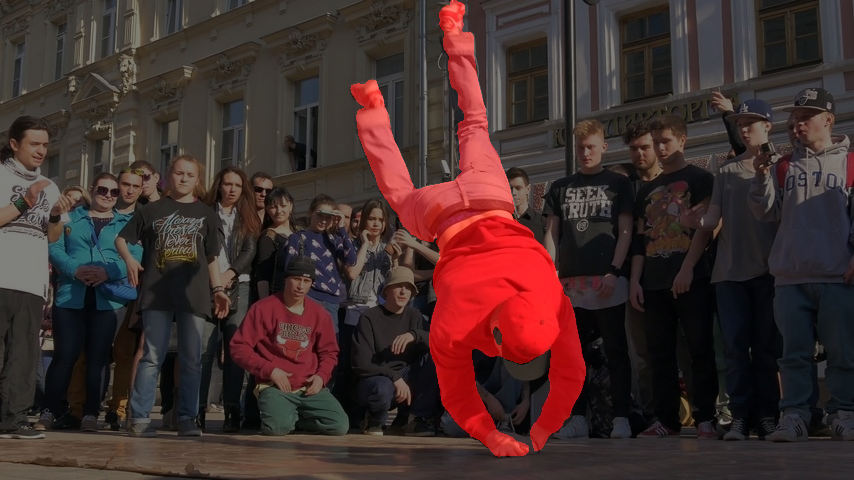} &
        \includegraphics[width=0.15\linewidth]{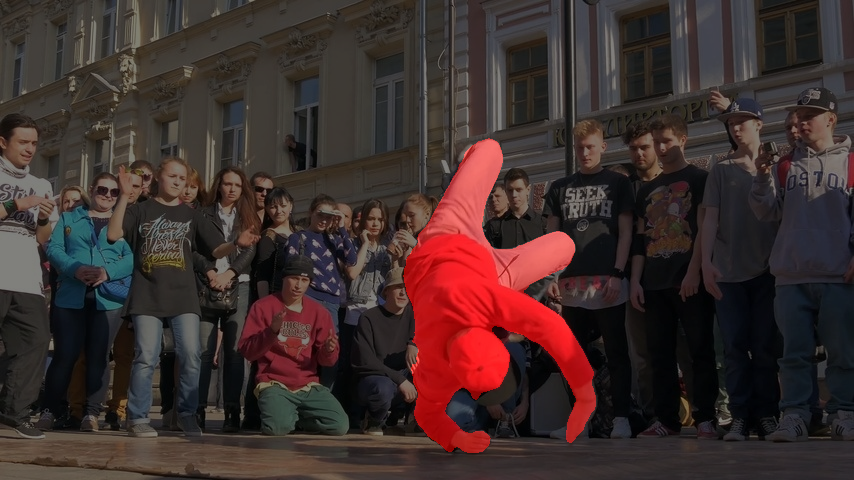} &
        \includegraphics[width=0.15\linewidth]{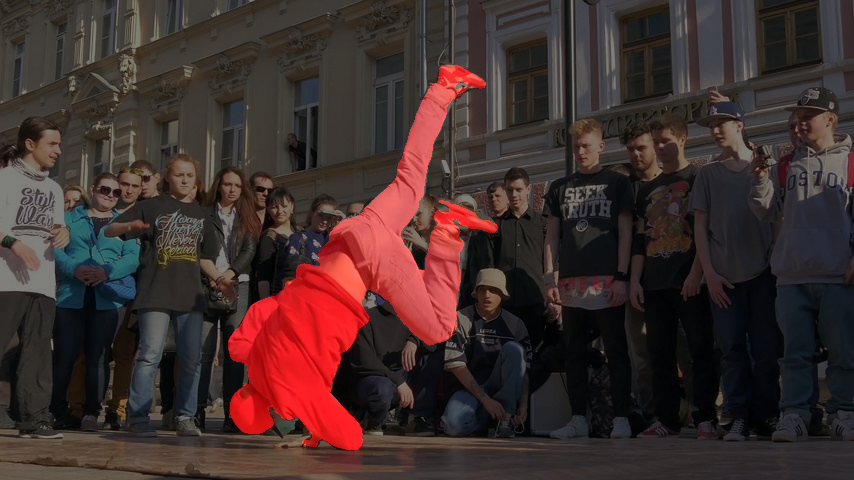} 
        \\ 
        \rotatebox{90}{\textbf{Case2}} &
        \includegraphics[width=0.15\linewidth]{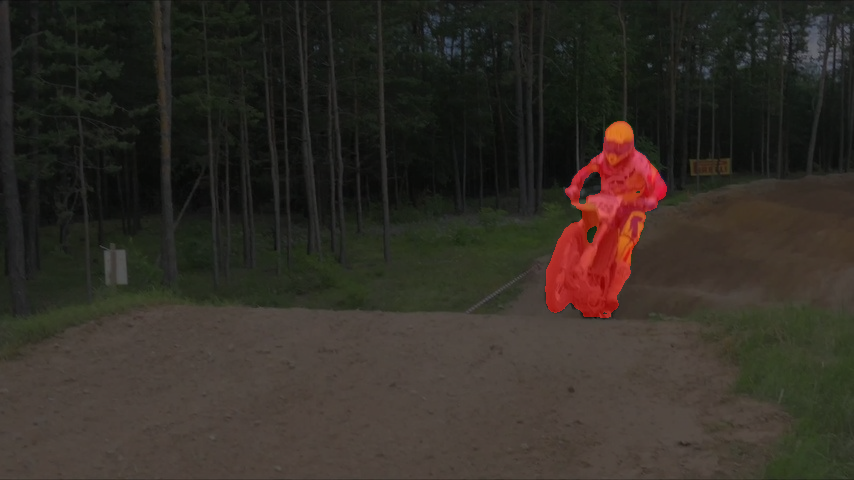} & 
        \includegraphics[width=0.15\linewidth]{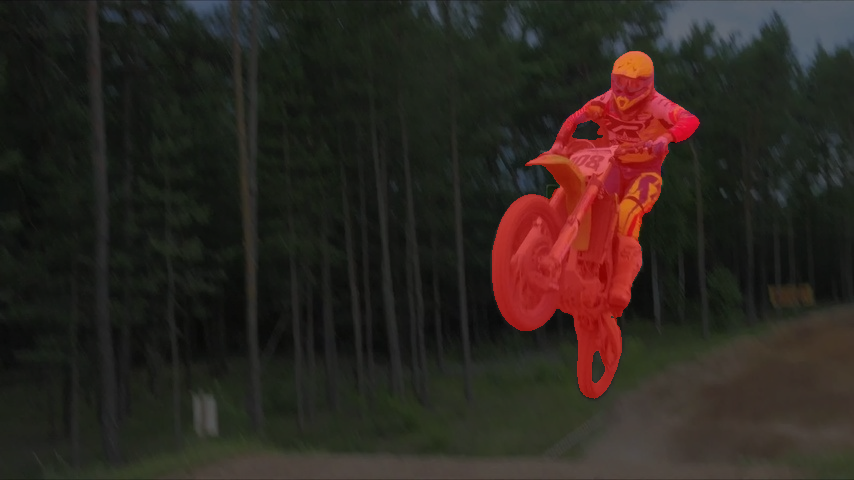} & 
        \includegraphics[width=0.15\linewidth]{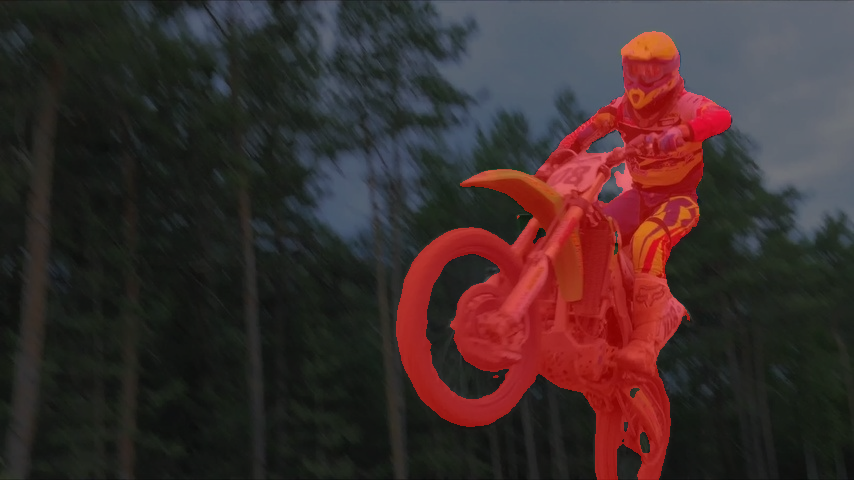} & 
        \includegraphics[width=0.15\linewidth]{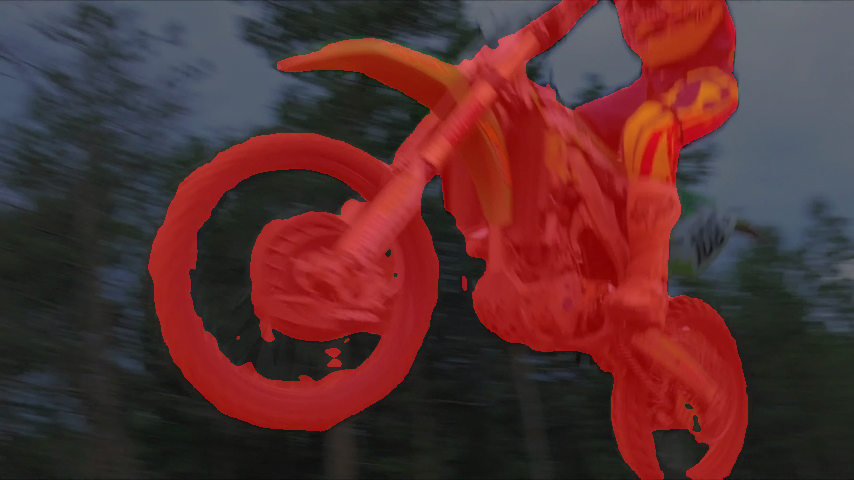} & 
        \includegraphics[width=0.15\linewidth]{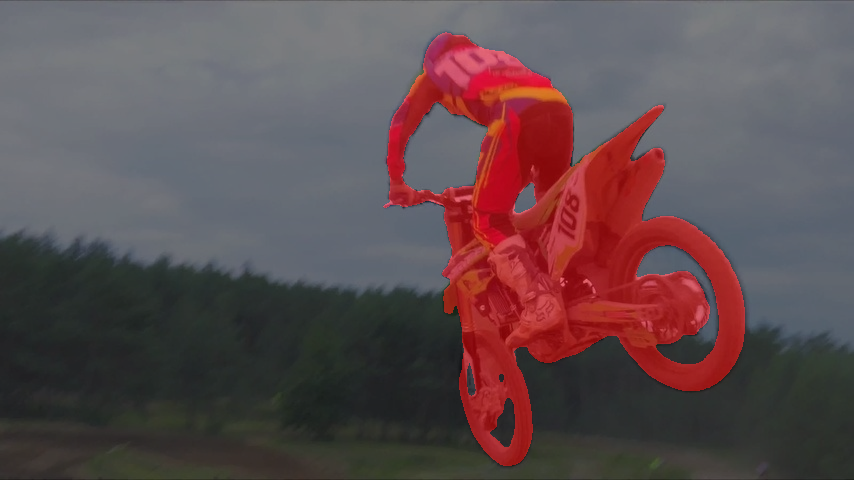} & 
        \includegraphics[width=0.15\linewidth]{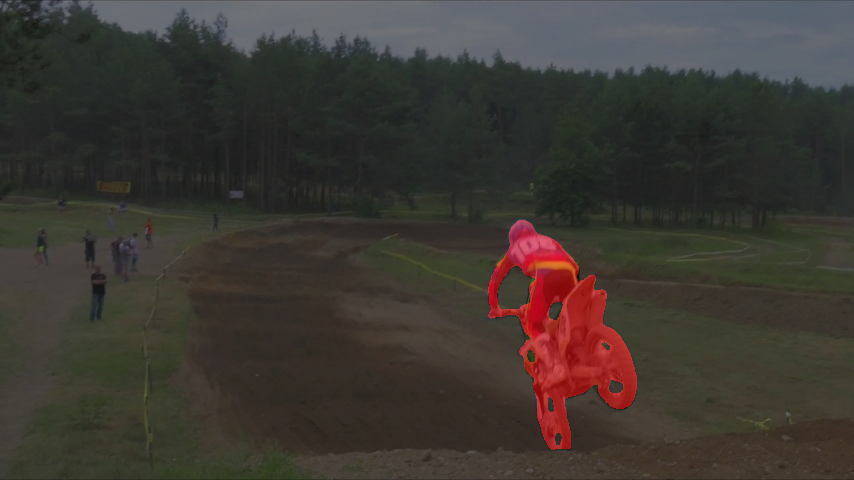} 
        \\  
        \rotatebox{90}{\textbf{Case3}} &
        \includegraphics[width=0.15\linewidth]{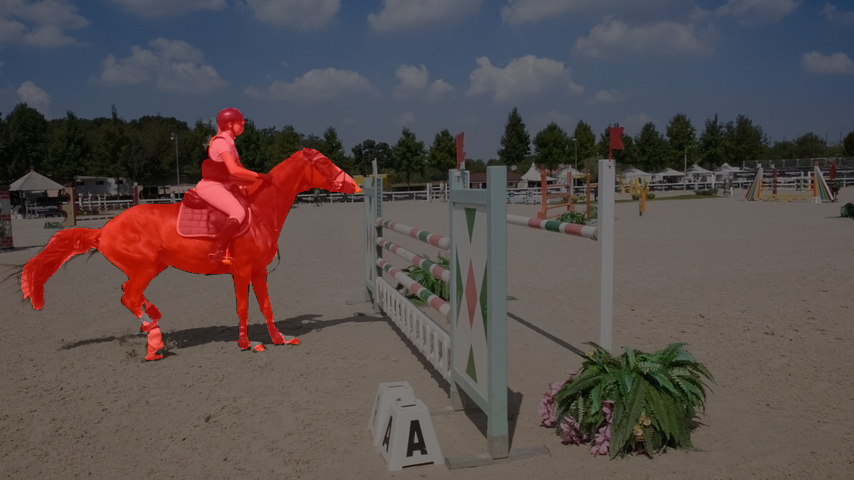} & 
        \includegraphics[width=0.15\linewidth]{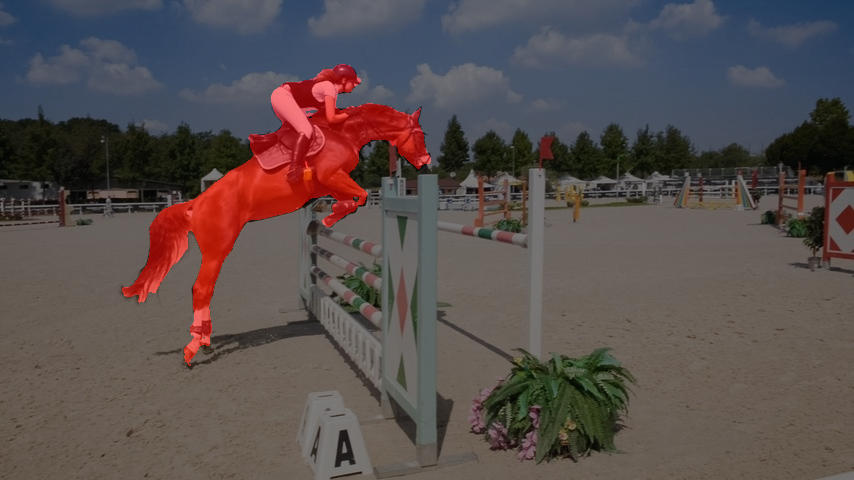} & 
        \includegraphics[width=0.15\linewidth]{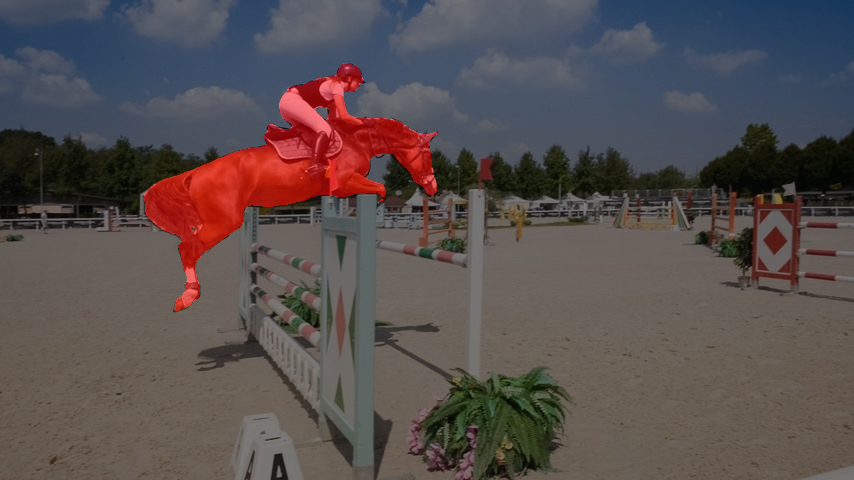} & 
        \includegraphics[width=0.15\linewidth]{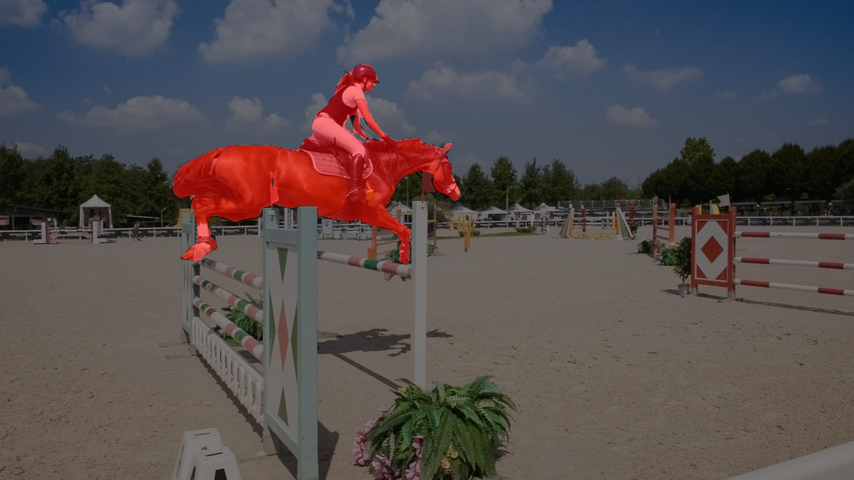} & 
        \includegraphics[width=0.15\linewidth]{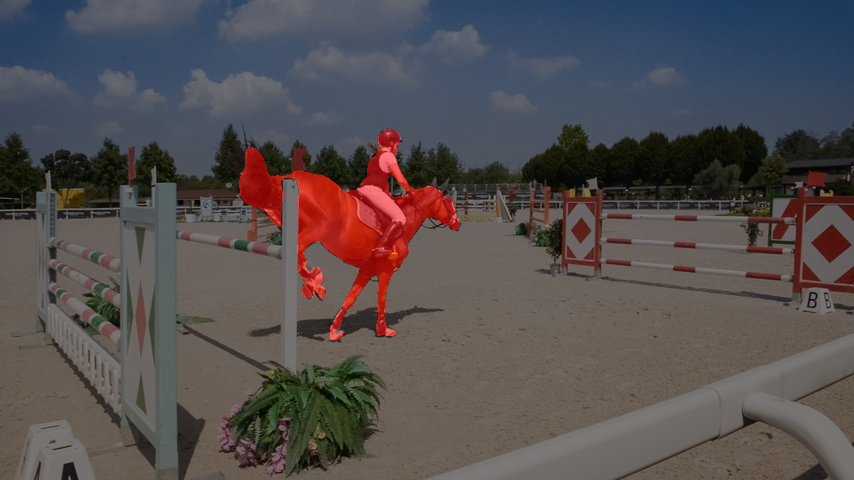} & 
        \includegraphics[width=0.15\linewidth]{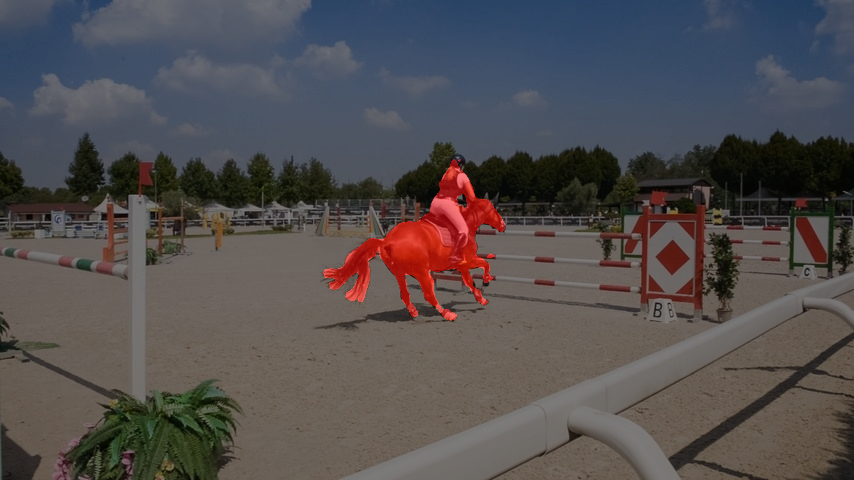} 
        \\  
        \rotatebox{90}{\textbf{Case4}} &
        \includegraphics[width=0.15\linewidth]{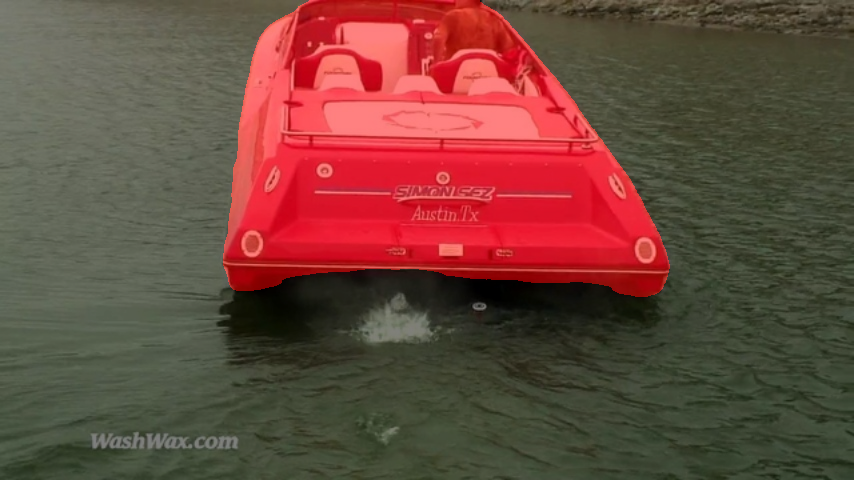} & 
        \includegraphics[width=0.15\linewidth]{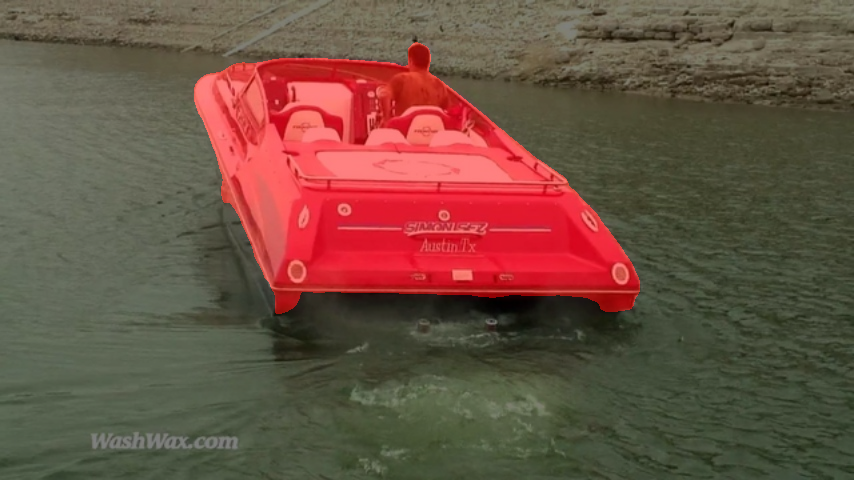} & 
        \includegraphics[width=0.15\linewidth]{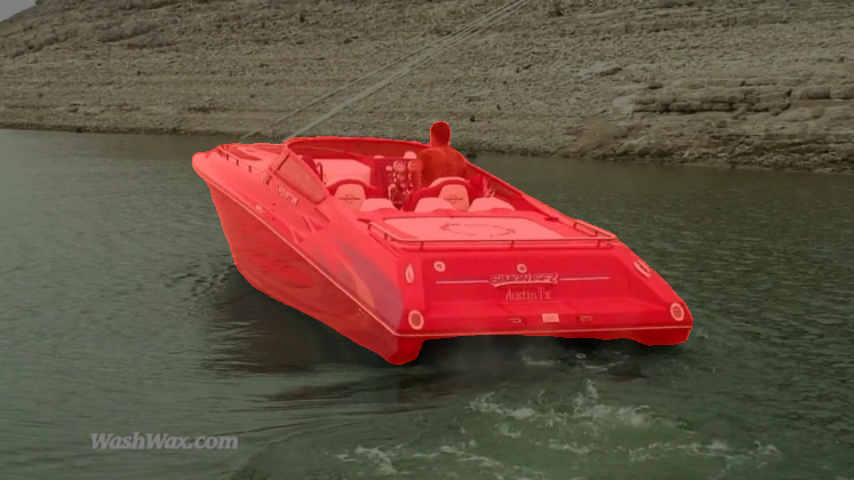} & 
        \includegraphics[width=0.15\linewidth]{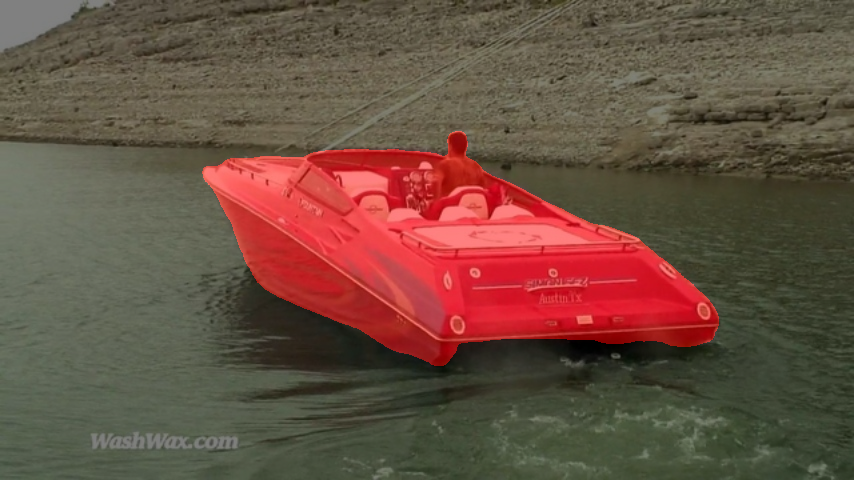} & 
        \includegraphics[width=0.15\linewidth]{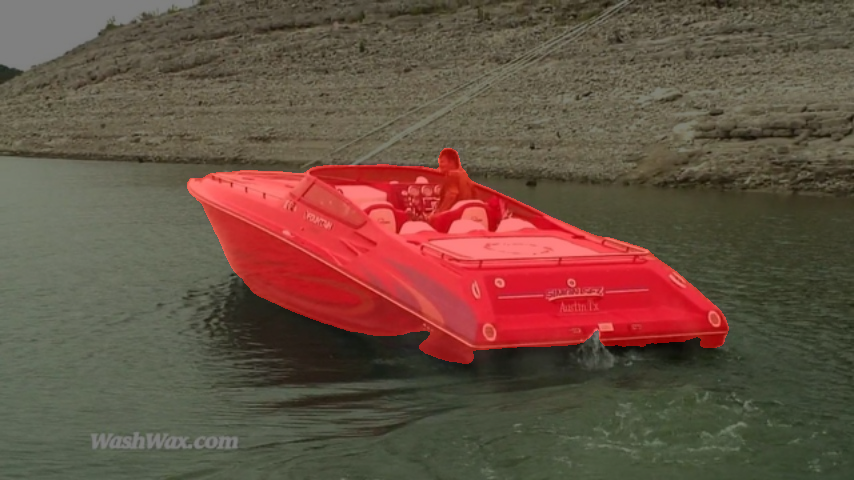} & 
        \includegraphics[width=0.15\linewidth]{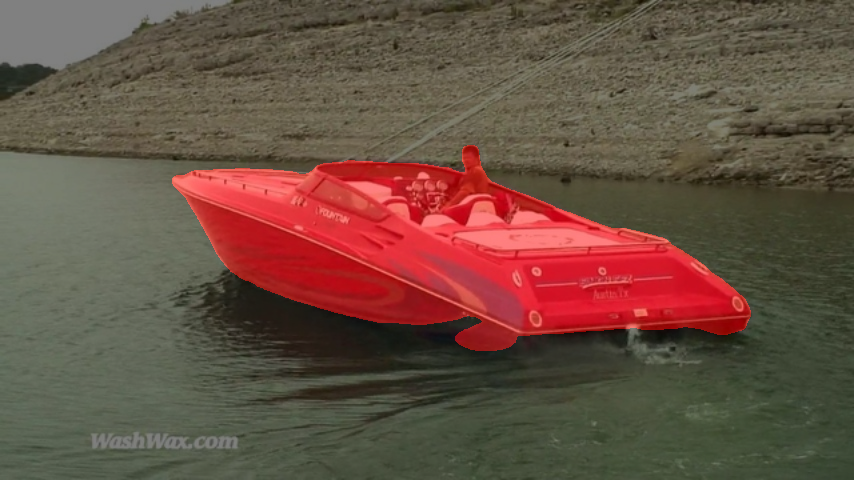} 
        \\ 
        \rotatebox{90}{\;\textbf{Case5}} &
        \includegraphics[width=0.15\linewidth]{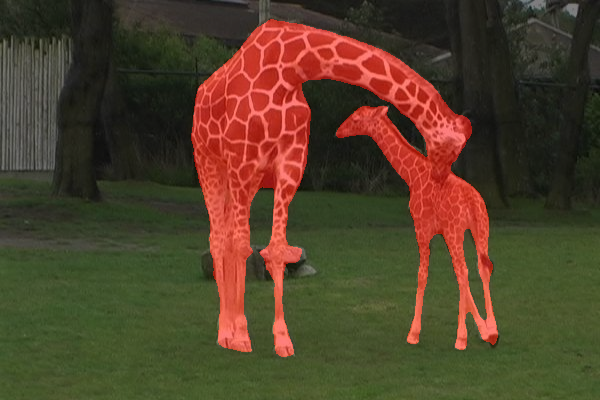} & 
        \includegraphics[width=0.15\linewidth]{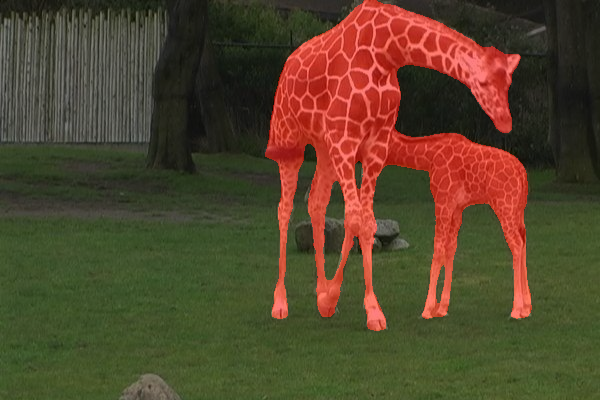} & 
        \includegraphics[width=0.15\linewidth]{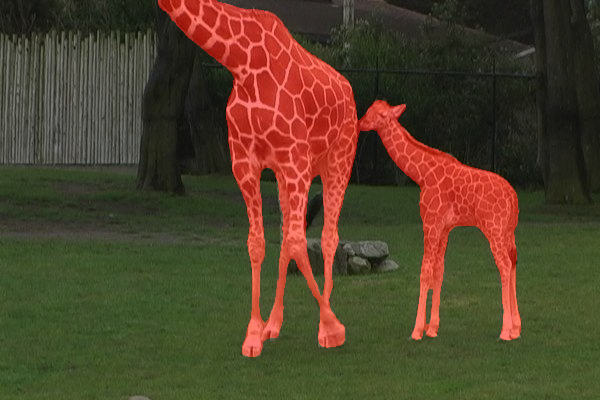} & 
        \includegraphics[width=0.15\linewidth]{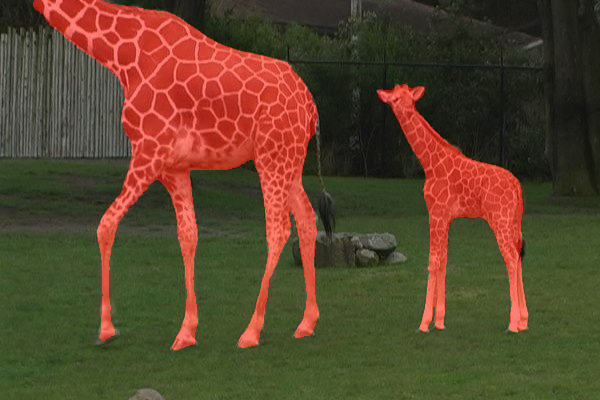} & 
        \includegraphics[width=0.15\linewidth]{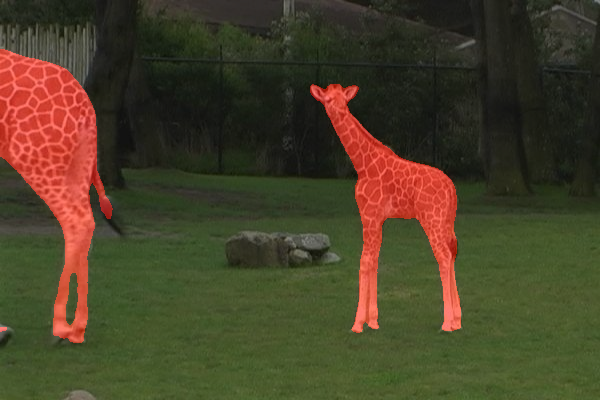} & 
        \includegraphics[width=0.15\linewidth]{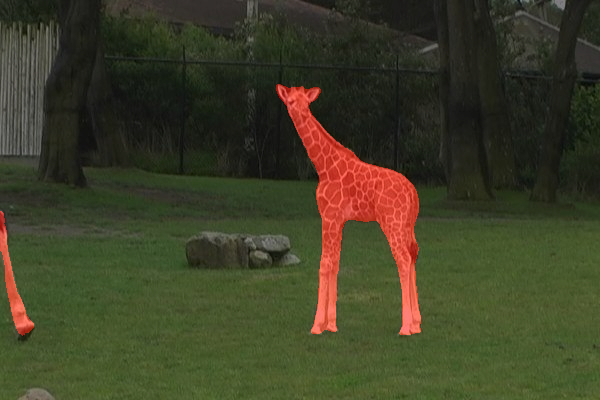} 
        \\ 
        \rotatebox{90}{\;\textbf{Case6}} &
        \includegraphics[width=0.15\linewidth]{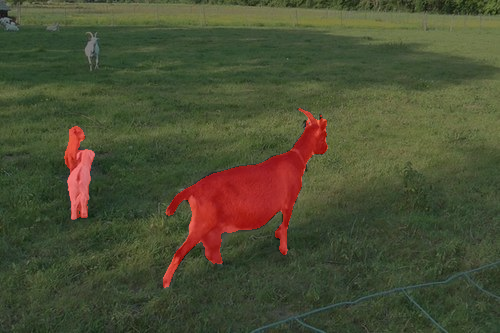} & 
        \includegraphics[width=0.15\linewidth]{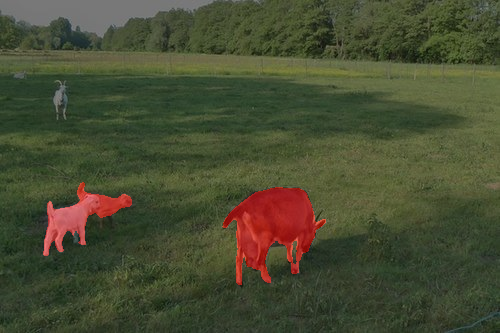} & 
        \includegraphics[width=0.15\linewidth]{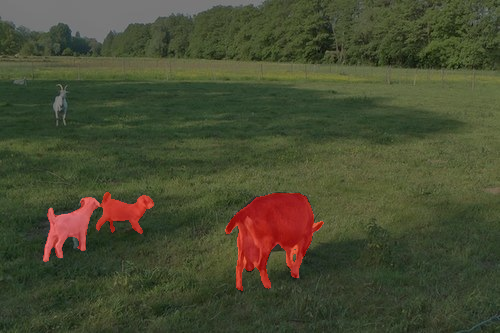} & 
        \includegraphics[width=0.15\linewidth]{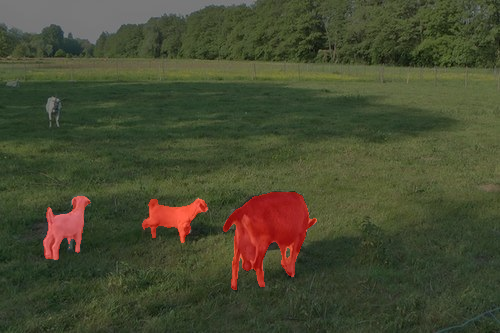} & 
        \includegraphics[width=0.15\linewidth]{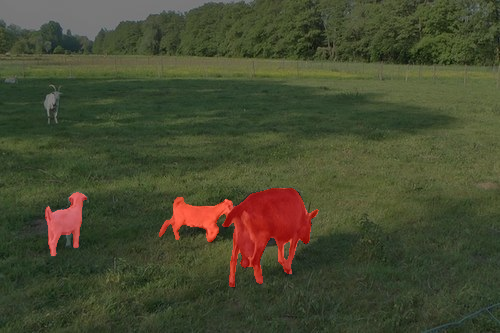} & 
        \includegraphics[width=0.15\linewidth]{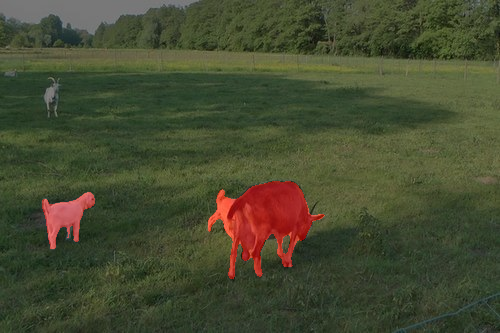} 
    \end{tabular}
    \caption{Qualitative visualization of segmentation results on multiple challenging scenarios, including rapid movement, fine-grained segmentation, motion blur, multiple objects, viewpoint variation, etc. Case 1-3: \textit{breakdance, motocross-jump, horsejump-high} from DAVIS-16; Case 4: \textit{boat0001\_00161} from YouTube-Obects; Case 5-6: \textit{giraffes01, goats01} from FBMS.}
    \label{Qualitative results UVOS}
\end{figure}

\begin{figure}[h]
    \centering
    \setlength{\tabcolsep}{1pt}
    \footnotesize
    \begin{tabular}{ccccccc}
         &  & \textbf{Case 1} &  &  & \textbf{Case 2} &  \\
        \rotatebox{90}{\textbf{Image}} &
        \includegraphics[width=0.15\linewidth, height=0.8cm]{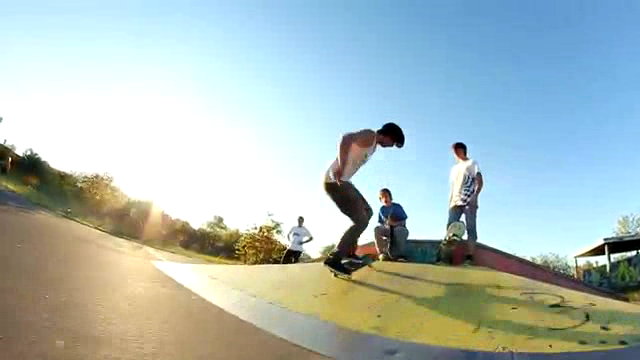} &
        \includegraphics[width=0.15\linewidth, height=0.8cm]{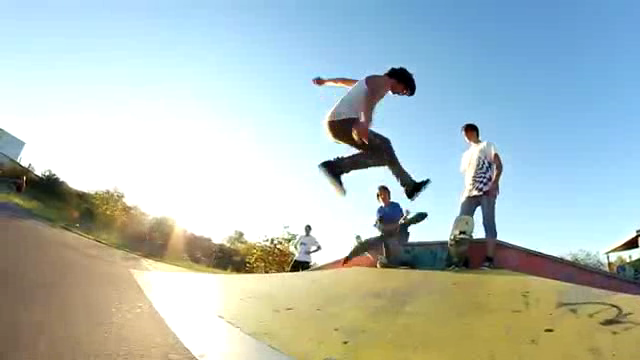} &
        \includegraphics[width=0.15\linewidth, height=0.8cm]{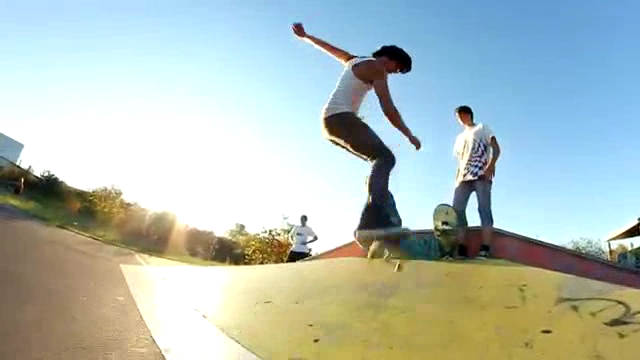}& 
        \includegraphics[width=0.15\linewidth, height=0.8cm]{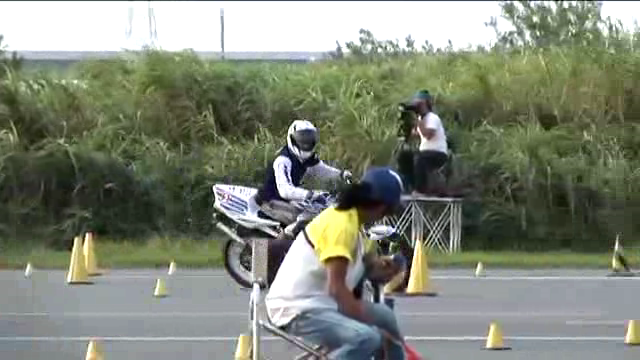} & 
        \includegraphics[width=0.15\linewidth, height=0.8cm]{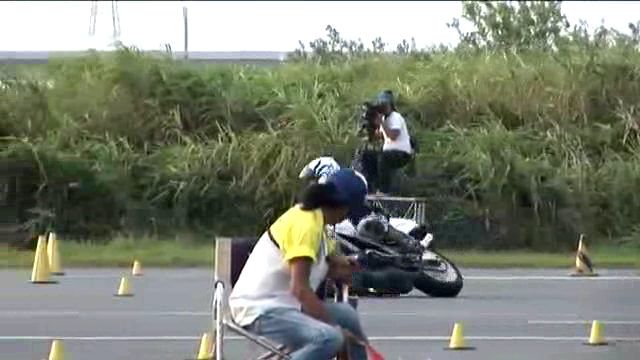} & 
        \includegraphics[width=0.15\linewidth, height=0.8cm]{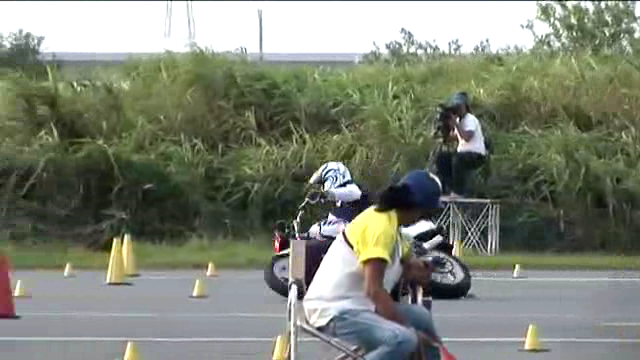}  \\

       \rotatebox{90}{\quad \textbf{GT}} &
        \includegraphics[width=0.15\linewidth, height=0.8cm]{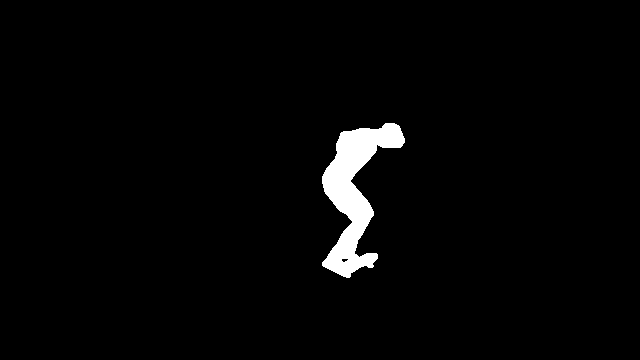} &
        \includegraphics[width=0.15\linewidth, height=0.8cm]{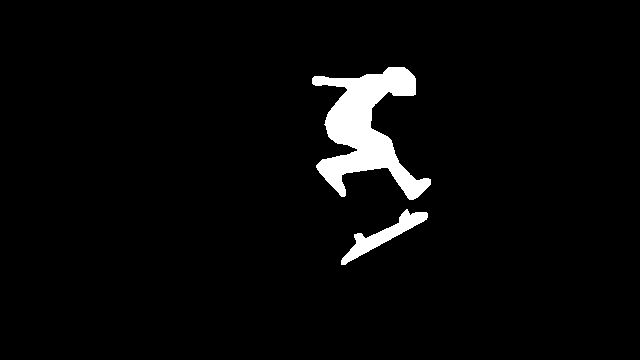} &
        \includegraphics[width=0.15\linewidth, height=0.8cm]{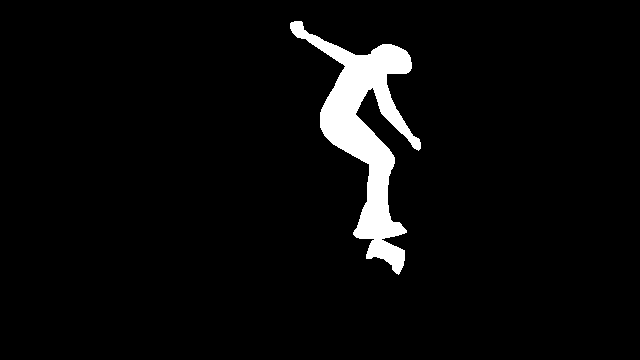}& 
        \includegraphics[width=0.15\linewidth, height=0.8cm]{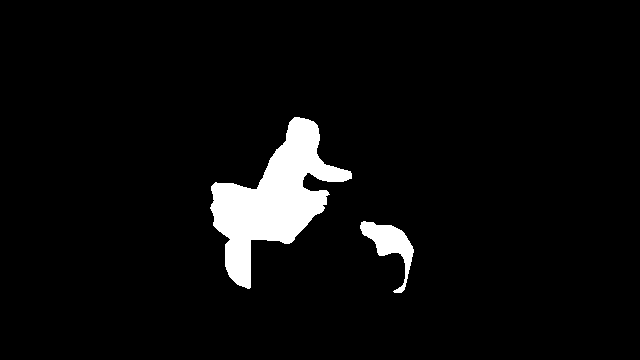} & 
        \includegraphics[width=0.15\linewidth, height=0.8cm]{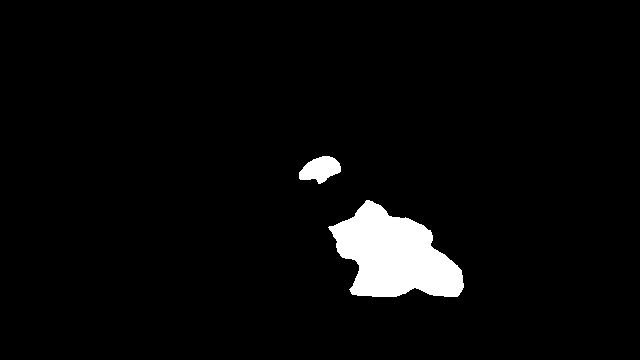} & 
        \includegraphics[width=0.15\linewidth, height=0.8cm]{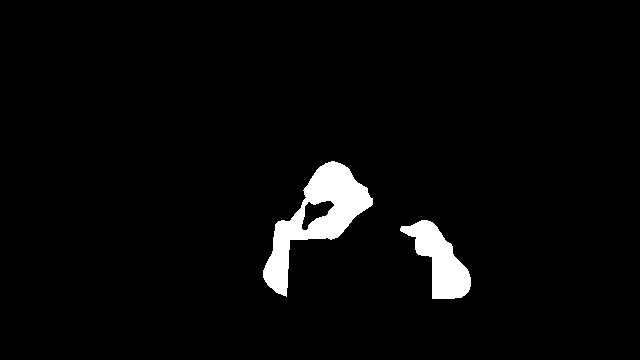}  \\

        \rotatebox{90}{\textbf{\scriptsize{MATNet}}} &
        \includegraphics[width=0.15\linewidth, height=0.8cm]{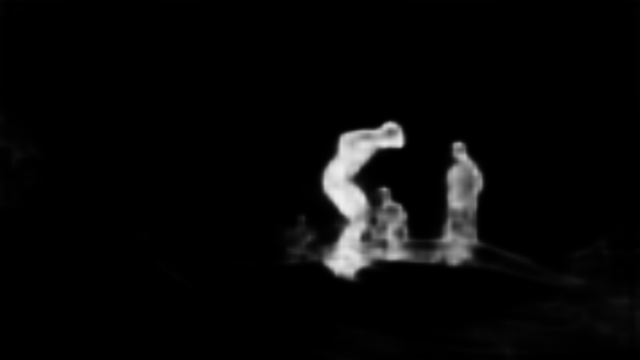} &
        \includegraphics[width=0.15\linewidth, height=0.8cm]{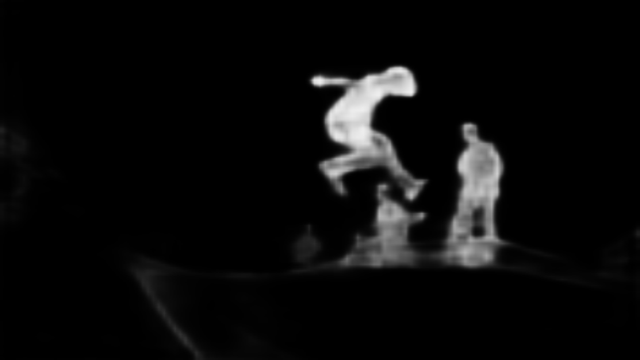} &
        \includegraphics[width=0.15\linewidth, height=0.8cm]{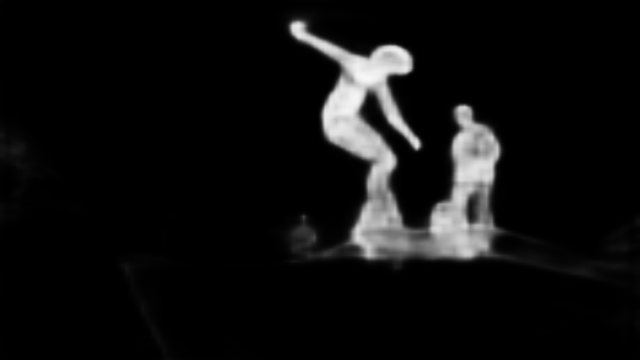}& 
        \includegraphics[width=0.15\linewidth, height=0.8cm]{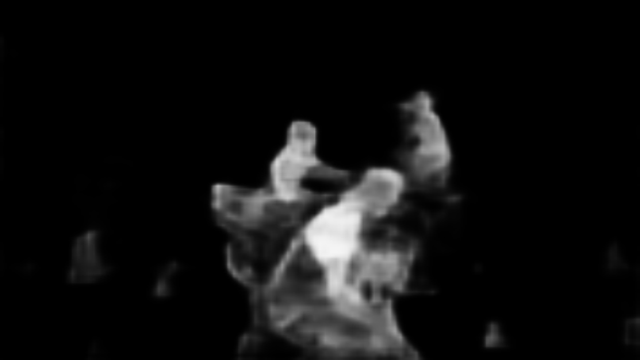} & 
        \includegraphics[width=0.15\linewidth, height=0.8cm]{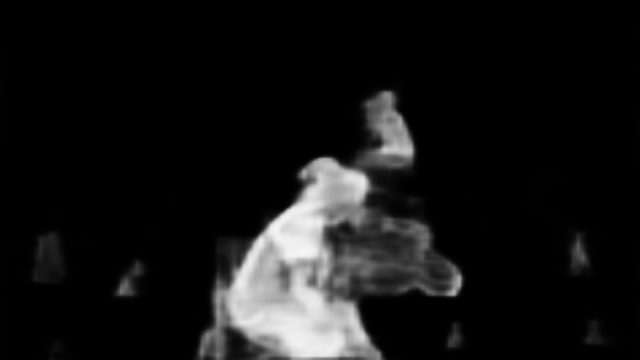} & 
        \includegraphics[width=0.15\linewidth, height=0.8cm]{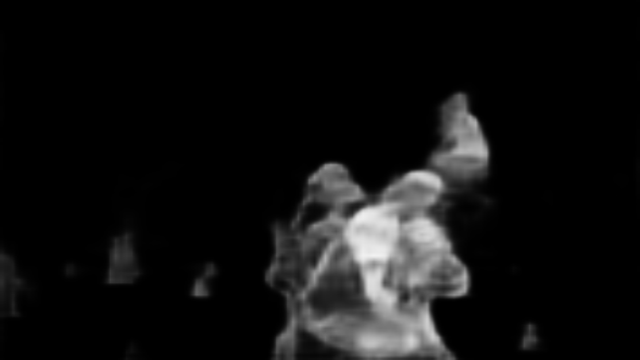}  \\

        \rotatebox{90}{\;\textbf{HFAN}} &
        \includegraphics[width=0.15\linewidth, height=0.8cm]{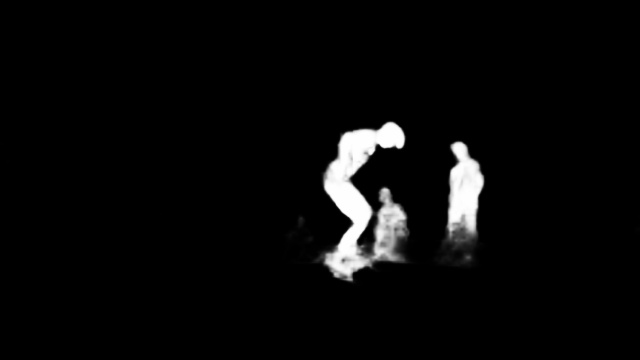} &
        \includegraphics[width=0.15\linewidth, height=0.8cm]{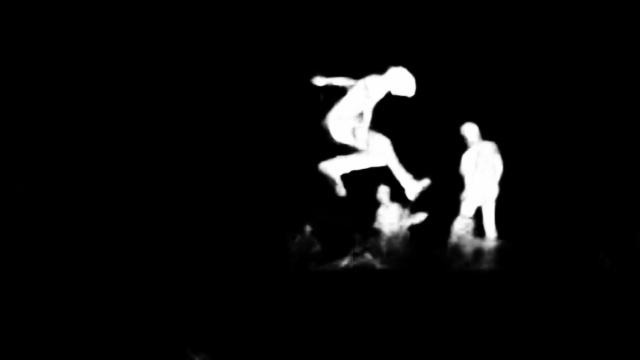} &
        \includegraphics[width=0.15\linewidth, height=0.8cm]{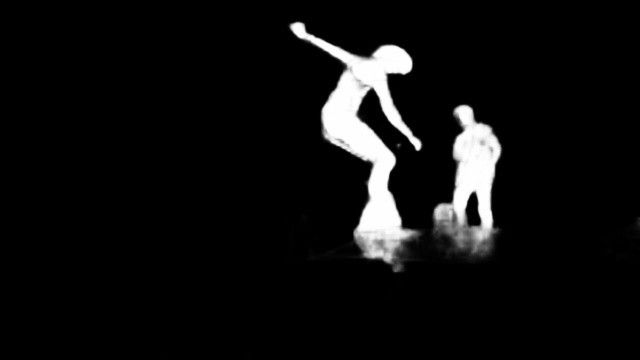}& 
        \includegraphics[width=0.15\linewidth, height=0.8cm]{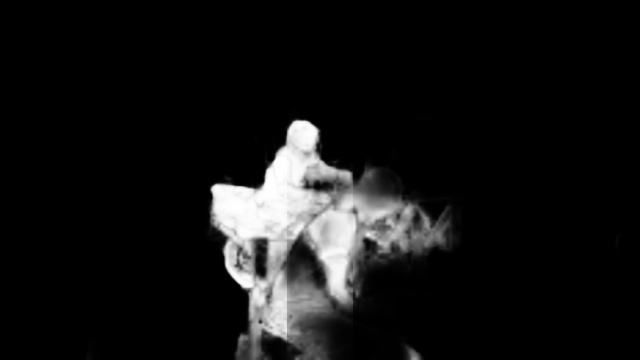} & 
        \includegraphics[width=0.15\linewidth, height=0.8cm]{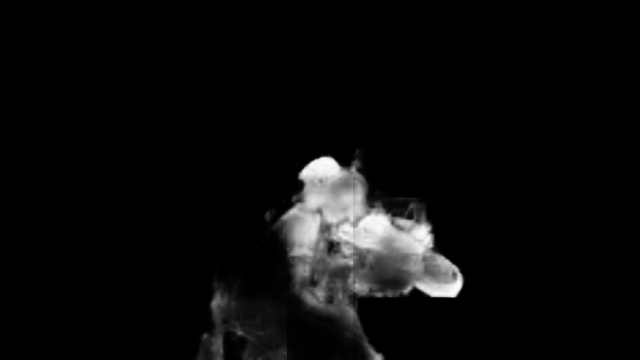} & 
        \includegraphics[width=0.15\linewidth, height=0.8cm]{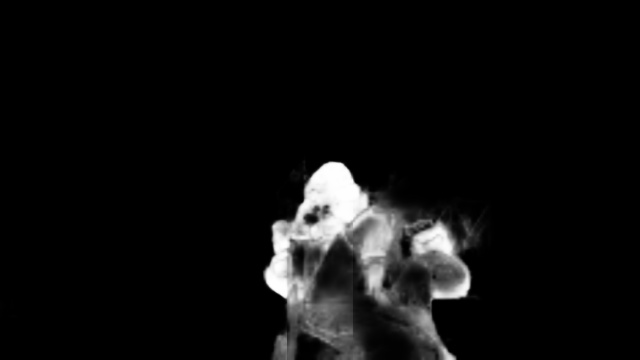}  \\

        \rotatebox{90}{\quad\textbf{TMO}} &
        \includegraphics[width=0.15\linewidth, height=0.8cm]{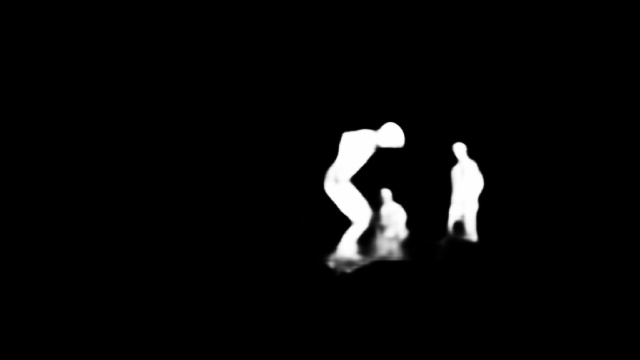} &
        \includegraphics[width=0.15\linewidth, height=0.8cm]{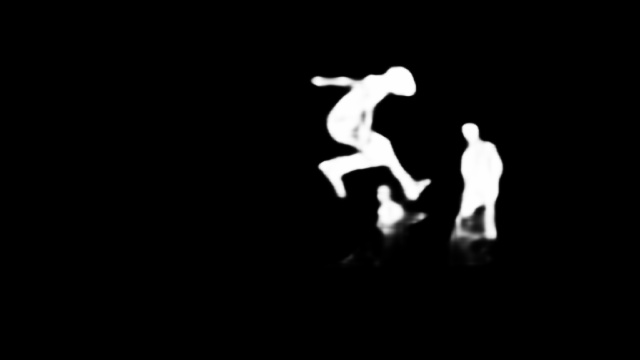} &
        \includegraphics[width=0.15\linewidth, height=0.8cm]{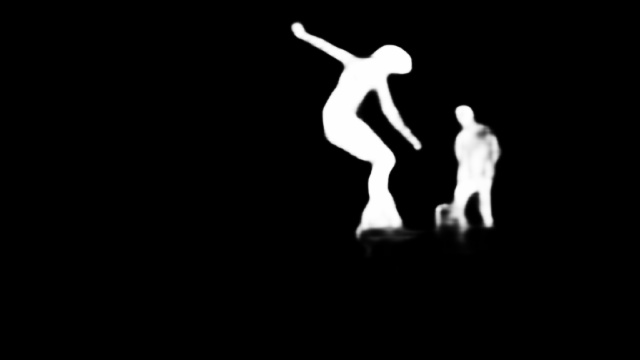}& 
        \includegraphics[width=0.15\linewidth, height=0.8cm]{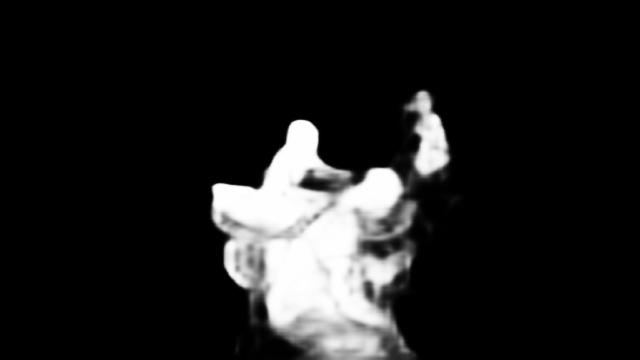} & 
        \includegraphics[width=0.15\linewidth, height=0.8cm]{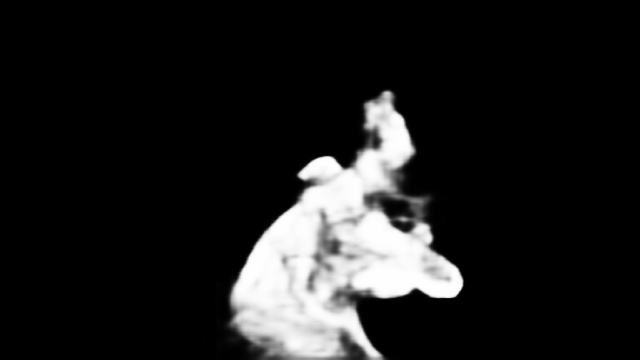} & 
        \includegraphics[width=0.15\linewidth, height=0.8cm]{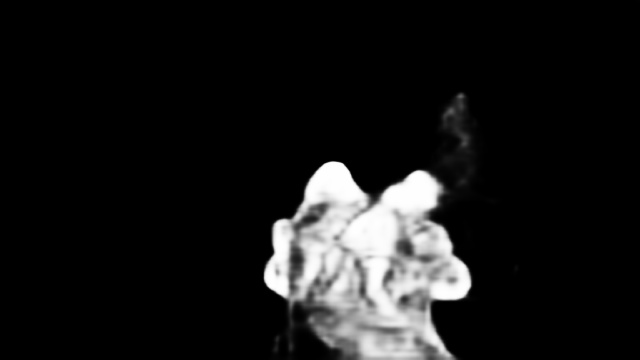}  \\

        \rotatebox{90}{\quad\textbf{Ours}} &
        \includegraphics[width=0.15\linewidth, height=0.8cm]{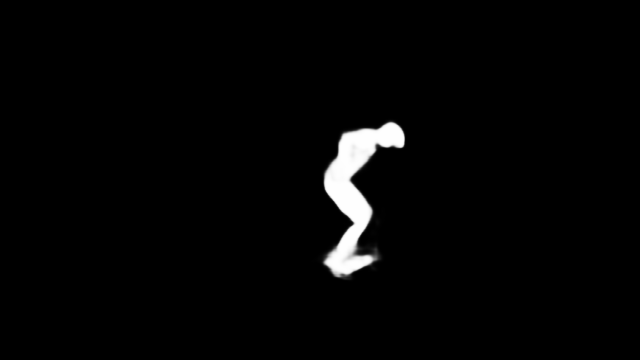} &
        \includegraphics[width=0.15\linewidth, height=0.8cm]{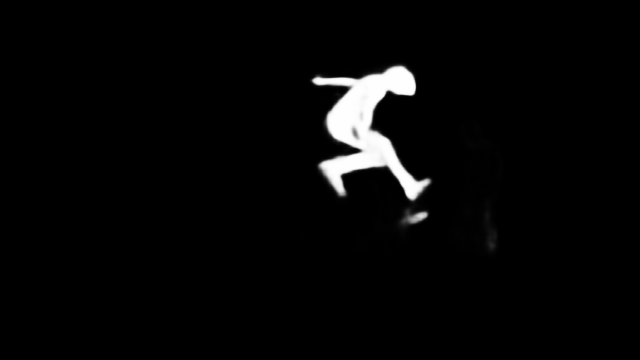} &
        \includegraphics[width=0.15\linewidth, height=0.8cm]{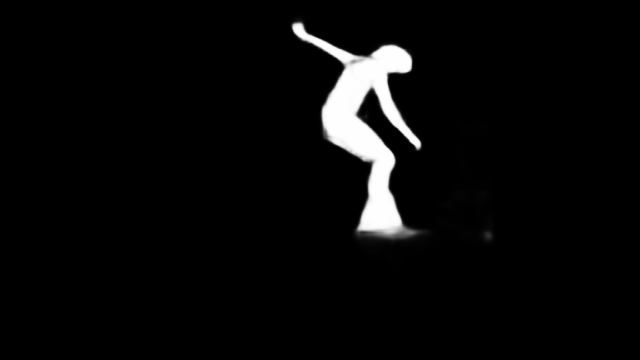}& 
        \includegraphics[width=0.15\linewidth, height=0.8cm]{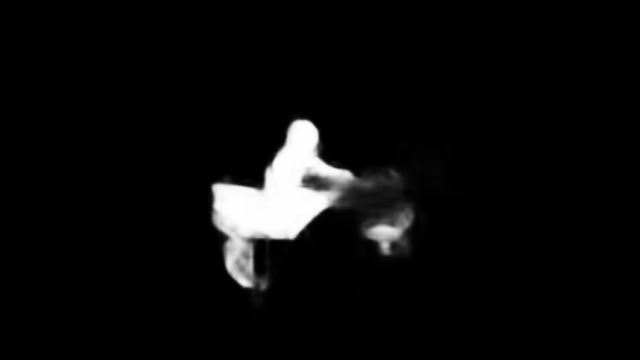} & 
        \includegraphics[width=0.15\linewidth, height=0.8cm]{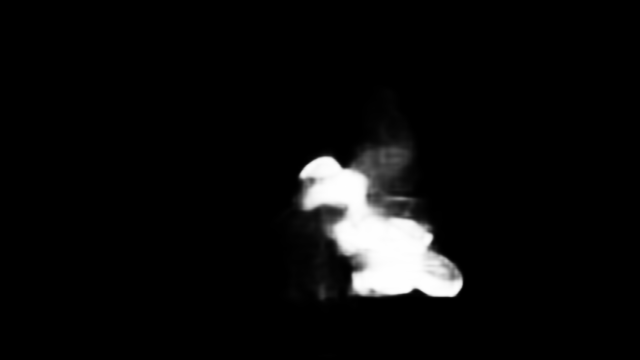} & 
        \includegraphics[width=0.15\linewidth, height=0.8cm]{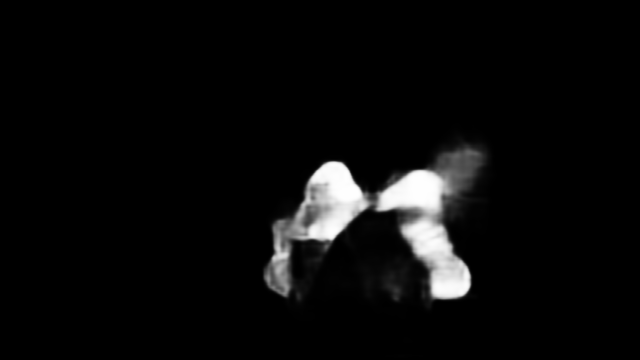}  \\
    \end{tabular}
    \caption{Visual demonstration and comparison of the UVOS models' performance on the challenging VSOD scenarios. Case 1-2: \textit{select\_019, select\_0194} from DAVSOD.}
    \label{Qualitative results VSOD}
\end{figure}

To intuitively demonstrate the capability of our model on both the UVOS and VSOD tasks, we provide qualitative visualizations of the segmentation results in several challenging scenarios in Fig. \ref{Qualitative results UVOS} and Fig. \ref{Qualitative results VSOD}. Our model delivers consistently accurate and complete detections across all challenging UVOS and VSOD cases, highlighting the model's robustness and generalization capability in handling diverse and challenging conditions.  

\subsection{Ablation Study}

\textbf{Impact of Memory Layer Selection}. We examined the impact of memory mechanism at different encoder layers on UVOS performance. As shown in Tab. \ref{tab:ablation_memory_layer}, incorporating memory at various levels consistently improves segmentation results. The memory applied at the second layer yields the best performance with an increase of 0.8\% \(\boldsymbol{\mathcal{J}} \& \boldsymbol{\mathcal{F}}\) on DAVIS-16, followed by that at the third layer with a 0.6\% \(\boldsymbol{\mathcal{J}} \& \boldsymbol{\mathcal{F}}\) increase. In contrast, memory at the final layer brings only marginal gains, supporting our theoretical claim that UVOS inherently lacks pixel-level supervision. Consequently, relying solely on high-level semantic memory bring about limited benefits for fine-grained object segmentation. We also tested their resource consumption, including the inference speed \textbf{Spd}, GPU memory usage \textbf{GPU} and the number of parameters \textbf{Pars}.

\begin{table}[h]
  \centering
  \caption{Ablation study on the impact of memory layer selection}
  \label{tab:ablation_memory_layer}
  \renewcommand{\arraystretch}{0.7}
   \begin{tabularx}{\linewidth}{l|p{1.15cm}p{1.15cm}p{1.15cm}p{0.5cm}p{0.55cm}p{0.55cm}}

    \toprule
    \multirow{2}{*}{\textbf{Variant}} 
    & DAVIS16 & FBMS & YTBOBJ & \textbf{Spd} & \textbf{GPU} & \textbf{Pars} \\
    & \(\boldsymbol{\mathcal{J}} \& \boldsymbol{\mathcal{F}}\) 
    & \(\boldsymbol{\mathcal{J}}\) 
    & \(\boldsymbol{\mathcal{J}}\) 
    & \textit{FPS}
    & \textit{MB}
    & \textit{M} \\
    \midrule
    baseline  & 88.4 & 84.7 & 75.1 & 34.4 & 140.0 & 36.7 \\
    layer = 1 & 88.9~\textbf{(+0.5)} & 86.0~\textbf{(+1.3)} & 75.7~\textbf{(+0.6)} & 9.2 & 141.5 & 37.1\\
    layer = 2 & 89.2~\textbf{(+0.8)} & 85.9~\textbf{(+1.2)} & 75.9~\textbf{(+0.8)} & 31.5 & 143.9 & 37.7  \\
    layer = 3 & 89.0~\textbf{(+0.6)} & 85.2~\textbf{(+0.5)} & 75.6~\textbf{(+0.5)} & 30.9 & 156.9 & 41.1\\
    layer = 4 & 88.6~\textbf{(+0.2)} & 84.9~\textbf{(+0.2)} & 76.1~\textbf{(+1.0)} & 31.8 & 178.0 & 46.7 \\
    \bottomrule
  \end{tabularx}
\end{table}

\textbf{Effectiveness of the Proposed Modules}. As shown in Tab. \ref{tab:ablation_module}, even without cross-memory feature interactions, the hierarchical memory architecture alone (denoted as Hier-Mem) already outperforms models with single-level memory. Introducing SGIM for high-to-shallow (abbreviated as H2S) and PLAM for shallow-to-high (abbreviated as S2H) interactions realizes the total performance improvements of 1.3\% and 1.0\% on DAVIS-16, respectively. And HMHI-Net achieves great improvements on all three UVOS bencmarks over the baseline model. To verify the heterogeneity of the interaction mechanisms, we swap SGIM and PLAM during shallow-high mutual refinement, which is denoted as 'Hetero' in Tab. \ref{tab:ablation_module}. And the degradation of performance in both cases confirms the necessity of our heterogeneous interaction mechanism. Additionally, we evaluated the efficiency of HMHI-Net and different modules on a single 4090 GPU, proving that HMHI-Net reaches the speed for reality application.

\begin{table}[h]
  \centering
  \caption{Ablation study on the effectiveness of the proposed modules}
  \label{tab:ablation_module}
  \renewcommand{\arraystretch}{0.8}
  \resizebox{\linewidth}{!}{
  \begin{tabular}{>{\centering\arraybackslash}cl|ccccc}
    \toprule
    \multicolumn{2}{c|}{\multirow{2}{*}{\textbf{Variant}}} 
    & DAVIS16 & FBMS & YTBOBJ & Speed & Params\\
    \multicolumn{2}{c|}{} 
    & \(\boldsymbol{\mathcal{J}} \& \boldsymbol{\mathcal{F}}\) 
    & \(\boldsymbol{\mathcal{J}}\) 
    & \(\boldsymbol{\mathcal{J}}\) 
    & \textit{FPS}
    & \textit{M}
    \\
    \midrule
    \multirow{5}{*}{\makecell{Modules}}
    & baseline             & 88.4 & 84.7 & 75.1 & 34.4 & 36.7 \\
    & \textit{w/} Multi-Mem          & 89.3~\textbf{(+0.9)} & 86.0~\textbf{(+1.3)} & 76.1~\textbf{(+1.0)} & 27.8 & 47.7\\
    & + S2H \textit{w/} PLAM           & 89.4~\textbf{(+1.0)} & 86.3~\textbf{(+1.6)} & 75.6~\textbf{(+0.5)} & 27.3 & 60.0\\
    & + H2S \textit{w/} SGIM           & 89.7~\textbf{(+1.3)} & 86.5~\textbf{(+1.8)} & 75.3~\textbf{(+0.2)} & 26.9 & 48.4 \\
    & HMHI-Net                & 89.8~\textbf{(+1.4)} & 86.9~\textbf{(+2.1)} & 76.2~\textbf{(+1.1)} & 26.2 & 60.8\\
    \midrule
    \multirow{2}{*}{\makecell{Hetero}}
    & S2H \textit{w/} SGIM          & 89.1~\textbf{(-0.3)} & 85.5~\textbf{(-0.8)}  & 73.9~\textbf{(-1.7)}  & - & -\\
    & H2S \textit{w/} PLAM          & 89.3~\textbf{(-0.4)}  & 84.8~\textbf{(-1.7)}  & 
    75.7~(+0.4) & - & -\\
    \bottomrule
  \end{tabular}}
\end{table}

\textbf{Evaluation of Model Robustness Across Backbones}. Finally, to assess the robustness of our proposed design, we integrate the hierarchical memory structure and heterogeneous interaction mechanism into various backbone architectures, including swin\_tiny from Swin-Transformer \cite{Swin-Transformer} and mit\_b1, mit\_b2, mit\_b3 from SegFormer \cite{Segformer}. Mark * denotes the use of the original backbone, yet others indicates modification following \cite{ISTC-Net}. Due to the large number of parameters in mit\_b2 and mit\_b3, baselines and HMHI-Net with these two backbones may not be fully trained. However, results presented in Tab. \ref{tab:ablation_robustness_backbone} still show consistent and significant performance improvements across all backbones, further validating the effectiveness and versatility of our approach.

\begin{table}[h]
  \centering
  \caption{Evaluation of model robustness across backbones}
  \renewcommand{\arraystretch}{0.7}
  \label{tab:ablation_robustness_backbone}
  \resizebox{\linewidth}{!}{
  \begin{tabular}{ll|ccc}
    \toprule
    \multicolumn{2}{c|}{\multirow{2}{*}{\textbf{Variant}}} 
    & DAVIS16 & FBMS & YTBOBJ\\
    \multicolumn{2}{c|}{} & \(\boldsymbol{\mathcal{J}} \& \boldsymbol{\mathcal{F}}\) & \(\boldsymbol{\mathcal{J}}\) & \(\boldsymbol{\mathcal{J}}\)\\
    \midrule
    \multirow{2}{*}{mit\_b1*} 
    & baseline & 87.8 & 82.5 & 75.3 \\
    & HMHI-Net   & 89.1~\textbf{(+1.3)} & 84.0~\textbf{(+3.0)} & 75.2\\
    \multirow{2}{*}{mit\_b2} 
    & baseline & 88.6 & 86.0 & 76 \\
    & HMHI-Net  & 89.6~\textbf{(+1.0)} & 86.5~\textbf{(+0.5)} & 75.7\\
    \multirow{2}{*}{mit\_b3} 
    & baseline & 88.0 & 86.4 & 76.4 \\
    & HMHI-Net   & 89.6~\textbf{(+1.6)} & 87.2~\textbf{(+0.8)} & 77.3~\textbf{(+0.9)}\\
    \multirow{2}{*}{swin\_tiny} 
    & baseline & 88.4 & 84.7 & 73.8 \\
    & HMHI-Net     & 89.4~\textbf{(+1.0)} & 85.5~\textbf{(+0.8)} & 76.1~\textbf{(+2.3)} \\
    \bottomrule
  \end{tabular}
  }
\end{table}

\begin{table}[h]
  \centering
  \caption{Ablation study on the influence of model inputs}
  \label{tab:ablation_model_input}
  \renewcommand{\arraystretch}{0.75}
  \resizebox{\linewidth}{!}{
  \begin{tabular}{cl|ccc}
    \toprule
    \multicolumn{2}{c|}{\multirow{2}{*}{\textbf{Variant}}} & DAVIS16 & FBMS & YTBOBJ\\
    \multicolumn{2}{c|}{} & \(\boldsymbol{\mathcal{J}} \& \boldsymbol{\mathcal{F}}\) & \(\boldsymbol{\mathcal{J}}\) & \(\boldsymbol{\mathcal{J}}\)\\
    \midrule
    \multirow{2}{*}{baseline}  
    & only\_flow & 78.8  & 63.8  & 60.0 \\
    & only\_image & 83.4  & 80.4  &  73.8\\
    \midrule
    \multirow{3}{*}{HMHI-Net} 
    & only\_flow & 82.3~\textbf{(+3.5)} & 66.8~\textbf{(+3.0)} & 62.1~\textbf{(+2.1)} \\
    & only\_image & 84.4~\textbf{(+1.0)} & 82.8~\textbf{(+2.4)} & 75.8~\textbf{(+2.0)} \\
    & flow \& image & 89.8 & 86.9 & 76.2\\
    \bottomrule
  \end{tabular}}
\end{table}

\textbf{Influence of Model Inputs}. We also studied the contribution of optical flow and RGB images as inputs. As shown in Tab. \ref{tab:ablation_model_input}, taking both features as inputs notably promotes model performance. Furthermore, even with single input, our model consistently surpasses the baseline under identical conditions with great margins. This validates HMHI-Net's capacity to effectively leverage video temporal cues to enhance segmentation.

\section{Conclusion}
We propose a simple and efficient hierarchical memory architecture with heterogeneous interaction mechanism for UVOS, which leverages both high-level features and shallow-level features for memory and performs different interactions during the shallow-high mutual refinement. Our model achieves state-of-the art performance on all UVOS and VSOD benchmarks. However, the hierarchical memory mechanism might lead to computation and storage overload, which can impact the model efficiency and is worth further investigation.

\section*{Acknowledgments}
This work was supported in part by CAAI-Lenovo Blue Sky Research Fund (No. CAAI-LXJJ 2024-05), and Scientific and Technological innovation action plan of Shanghai Science and Technology Committee (No.22511101502).

\newpage
\bibliographystyle{ACM-Reference-Format}
\nocite{*}
\bibliography{sample-sigconf-authordraft}
\end{document}